\pdfoutput=1
\RequirePackage{etoolbox}
\PassOptionsToPackage{disable}{todonotes}

\AtBeginDocument{%
  \iclrfinalcopy
  \author{%
    \parbox{\dimexpr\textwidth-2\tabcolsep\relax}{\centering
      Riccardo Zaccone\textsuperscript{1,3}\thanks{Equal contribution.} \quad 
      Avi Trost\textsuperscript{2}\footnotemark[1] \quad 
      Mart\'in Bravo\textsuperscript{3}\\
      Xinyang Xu\textsuperscript{2} \quad 
      Alexander Yun\textsuperscript{2} \quad 
      Frederic Sala\textsuperscript{2}\\
      Carlo Masone\textsuperscript{1} \quad 
      Stefanos Laskaridis\textsuperscript{4}{\thanks{Work done independently of Amazon.}} \quad 
      Samuel Horv\'ath\textsuperscript{3}\\[0.5em]
      \normalfont\small
      \textsuperscript{1}Polytechnic Institute of Turin\\
      \textsuperscript{2}University of Wisconsin--Madison\\
      \textsuperscript{3}Mohamed bin Zayed University of Artificial Intelligence\\
      \textsuperscript{4}Amazon Science\\[0.5em]
      Correspondence to: \texttt{riccardo.zaccone@mbzuai.ac.ae}
    }%
  }
  \colorlet{revisioncolor}{black}
  \renewcommand{\authorNote}[3]{}
  \renewcommand{\authorNoteMargin}[3]{}
  \renewcommand{\aviReply}[1]{}
  \renewcommand{\headrulewidth}{0pt}
  \apptocmd{\maketitle}{\fancyhead{}}{}{%
    \PackageError{arxiv}{Could not remove the conference header}{}
  }
  \hypersetup{
    hidelinks,
    pdftitle={Prefix-LM: Prefix-Invariant Transformers for Width-Adaptive Progressive Inference},
    pdfauthor={Riccardo Zaccone, Avi Trost, Mart\'in Bravo, Xinyang Xu, Alexander Yun, Frederic Sala, Carlo Masone, Stefanos Laskaridis, Samuel Horv\'ath}
  }
}

\PassOptionsToPackage{dvipsnames}{xcolor}
\documentclass{article} %
\usepackage{iclr2027_conference,times}

\usepackage{amsmath,amsfonts,bm}

\def\eqref#1{equation~\ref{#1}}

\def\1{\bm{1}}

\def\vs{{\bm{s}}}

\DeclareMathAlphabet{\mathsfit}{\encodingdefault}{\sfdefault}{m}{sl}
\SetMathAlphabet{\mathsfit}{bold}{\encodingdefault}{\sfdefault}{bx}{n}

\usepackage{hyperref}
\usepackage{url}
\usepackage{graphicx}
\usepackage{booktabs}
\usepackage[dvipsnames]{xcolor}
\usepackage[dvipsnames]{xcolor}
\usepackage{xcolor-material, colortbl, soul, wrapfig, needspace}
\usepackage{nicefrac, xspace, multirow, makecell, sidenotes}

\usepackage{amsmath}
\usepackage{amssymb}
\usepackage{mathtools}
\usepackage{amsthm}
\usepackage{thmtools} %
\usepackage[capitalize,noabbrev]{cleveref}
\usepackage{xspace}
\usepackage{array}
\usepackage{enumitem}
\newcolumntype{C}[1]{>{\centering\arraybackslash}p{#1}}

\usepackage{makecell}

\usepackage{pifont}%

\usepackage[
  colorinlistoftodos,
  textsize=scriptsize
]{todonotes}

\usepackage{marginnote}
\usepackage{tikz}

\newcommand{\authorNote}[3]{%
  \todo[
    inline,
    bordercolor=#2,
    backgroundcolor=#2!20,
    linecolor=#2
  ]{\textbf{#1: }#3}%
}

\newcounter{notecount}

\newcommand{\authorNoteMargin}[3]{%
  \stepcounter{notecount}%
  \tikz[remember picture,baseline,overlay]
    \node[inner sep=0pt] (note-origin-\thenotecount) {};%
  \marginnote{%
    \tikz[remember picture,overlay]
      \node[
        anchor=west,
        inner sep=0pt,
        text width=\marginparwidth
      ] (note-box-\thenotecount) {%
        \colorbox{#2!20}{%
          \parbox{\dimexpr\marginparwidth-2\fboxsep\relax}{%
            \normalfont\scriptsize
            \textbf{#1: }#3%
          }%
        }%
      };%
    \tikz[remember picture,overlay]
      \draw[#2]
        (note-origin-\thenotecount)
        -- (note-box-\thenotecount.west);%
  }%
}

\newcommand{\aviReply}[1]{%
  \par\smallskip
  \colorbox{green!20}{%
    \parbox{\dimexpr\marginparwidth-2\fboxsep\relax}{%
      \normalfont\scriptsize
      \textbf{Avi: }#1%
    }%
  }%
}

\definecolor{revisioncolor}{RGB}{0,70,160}
\DeclareCaptionFont{revision}{\color{revisioncolor}}

\newif\ifrevisionactive

\newenvironment{revised}
  {\par\begingroup
   \revisionactivetrue
   \color{revisioncolor}%
   \captionsetup{
     font+=revision,
     labelfont+=revision,
     textfont+=revision}}
  {\par\endgroup}

\makeatletter
\apptocmd{\@floatboxreset}{%
  \ifrevisionactive
    \color{revisioncolor}%
  \fi
}{}{\PackageWarning{revision}{Could not patch float color}}
\makeatother

\definecolor{navyblue}{RGB}{0,35,90}

\theoremstyle{plain}
\newtheorem{theorem}{Theorem}[section]
\newtheorem{proposition}[theorem]{Proposition}
\newtheorem{lemma}[theorem]{Lemma}

\newtheorem{corollary}[theorem]{Corollary}
\theoremstyle{definition}
\newtheorem{definition}[theorem]{Definition}

\theoremstyle{remark}

\definecolor{Gray}{gray}{0.9}
\usepackage{algorithm}
\usepackage{algorithmic}
\usepackage{tcolorbox}

\makeatletter
\DeclareRobustCommand\onedot{\futurelet\@let@token\@onedot}
\def\@onedot{\ifx\@let@token.\else.\null\fi\xspace}

 \def\vs{\emph{vs}\onedot}

\makeatother

\makeatletter
 \def\SOUL@hlpreamble{%
 \setul{}{3.2ex}%
 \let\SOUL@stcolor\SOUL@hlcolor
 \SOUL@stpreamble
 }
\makeatother

\DeclareRobustCommand{\gpt}
{\textsc{GPT-2}}
\DeclareRobustCommand{\llama}
{\textsc{Llama3.2-1B}}
\DeclareRobustCommand{\nemotron}
{\textsc{Nemotron-Flash-$1$B}}
\DeclareRobustCommand{\qwen}
{\textsc{Qwen3.5-$2$B}}

\crefname{equation}{Eq.}{Eq.}
\Crefname{equation}{Equation}{Equations}
\creflabelformat{equation}{(#2#1#3)}

\crefname{figure}{Fig.}{Figs.}
\Crefname{figure}{Figure}{Figures}

\crefname{table}{Tab.}{Tabs.}
\Crefname{table}{Table}{Tables}

\crefname{chapter}{Chap.}{Chaps.}
\Crefname{chapter}{Chapter}{Chapters}

\crefname{section}{Sec.}{Secs.}
\Crefname{section}{Section}{Sections}

\crefname{subsection}{Sec.}{Secs.}
\Crefname{subsection}{Section}{Sections}

\crefname{subsubsection}{Sec.}{Secs.}
\Crefname{subsubsection}{Section}{Sections}

\crefname{prob}{Prob.}{Probs.}
\Crefname{prob}{Problem}{Problems}

\crefname{property}{Prop.}{Props.}
\Crefname{property}{Property}{Properties}

\crefname{constr}{Constraint}{Constraints} %
\Crefname{constr}{Constraint}{Constraints}

\crefname{algocf}{Algorithm}{Algorithms}
\Crefname{algocf}{Algorithm}{Algorithms}

\crefname{algorithm}{Algorithm}{Algorithms}
\Crefname{algorithm}{Algorithm}{Algorithms}

\crefname{hyp}{Hyp.}{Hyps.}
\Crefname{hyp}{Hypothesis}{Hypothesis}

\crefname{assumption}{Assumption}{Assumptions}
\Crefname{assumption}{Assumption}{Assumptions}

\crefname{lem}{Lem.}{Lems.}
\Crefname{lem}{Lemma}{Lemma}

\crefname{prop}{Prop.}{Props.}
\Crefname{prop}{Proposition}{Propositions}

\crefname{theorem}{Thm.}{Thms.}
\Crefname{theorem}{Thm.}{Thms.}

\crefname{definition}{Def.}{Defs.}
\Crefname{definition}{Def.}{Defs.}

\crefname{appendix}{App.}{Apps.}
\Crefname{appendix}{App.}{Apps.}

\newcommand{\method}{\textsc{Prefix-LM}\xspace}

\newcommand{\loss}{\mathcal{L}}
\newcommand{\hsd}{\textsc{HSD}\xspace}
\newcommand{\mng}{\textsc{MNG}\xspace}

\usepackage{pgfplots}
\usepgfplotslibrary{groupplots,colormaps}
\pgfplotsset{compat=1.18}

\newcommand{\PlotFont}{\scriptsize}
\newcommand{\PlotHeight}{1.8cm}
\newcommand{\PlotGap}{0.75cm}

\newcommand{\PlotLegendGap}{0.85cm}
\newcommand{\PlotLineWidth}{0.8pt}
\newcommand{\PlotMarkSize}{1.5pt}
\pgfplotsset{
  paper axis/.style={
    height=\PlotHeight,scale only axis,
    tick label style={font=\PlotFont},
    label style={font=\PlotFont},title style={font=\PlotFont\bfseries},
    grid=major,grid style={gray!20,line width=0.3pt},
    axis line style={gray!60,line width=0.4pt},
    tick style={gray!60,line width=0.4pt},
    tick align=outside,major tick length=2pt,
    every axis plot/.append style={line width=\PlotLineWidth,mark size=\PlotMarkSize},
    legend style={font=\PlotFont,draw=none,fill=none,cells={anchor=west},
      at={(0.5,0)},anchor=north,yshift=-\PlotLegendGap}
  },
  paper pair/.style={paper axis,width=0.37\linewidth},
  paper triple/.style={paper axis,width=\dimexpr(\linewidth-3.6cm)/3\relax},
  paper single/.style={paper axis,width=0.5\linewidth}
}

\title{\method{}: Prefix-Invariant Transformers \\ for Width-Adaptive Progressive Inference}

\author{Antiquus S.~Hippocampus, Natalia Cerebro \& Amelie P. Amygdale \thanks{ Use footnote for providing further information
about author (webpage, alternative address)---\emph{not} for acknowledging
funding agencies.  Funding acknowledgements go at the end of the paper.} \\
Department of Computer Science\\
Cranberry-Lemon University\\
Pittsburgh, PA 15213, USA \\
\texttt{\{hippo,brain,jen\}@cs.cranberry-lemon.edu} \\
\And
Ji Q. Ren \& Yevgeny LeNet \\
Department of Computational Neuroscience \\
University of the Witwatersrand \\
Joburg, South Africa \\
\texttt{\{robot,net\}@wits.ac.za} \\
\AND
Coauthor \\
Affiliation \\
Address \\
\texttt{email}
}

\begin{document}

\maketitle

\begin{abstract}
LLMs require a significant amount of resources during autoregressive generation, but not all tokens are equal.
Token prediction is a task of varying difficulty, but monolithic architectures allocate the same compute regardless of the difficulty of prediction.
Progressive inference, where representations are incrementally computed and refined as needed, can address this challenge.
This has traditionally been achieved through early-exiting methods that operate along the depth dimension of the network.
Elastic networks that operate along the width dimension can select a different computational budget per input, but the resulting outputs cannot be progressively refined.
We identify the missing property as \emph{prefix invariance}: the intermediate representations of smaller submodels are prefixes of larger ones.
To this end, we propose \method{}: an architectural design that achieves prefix invariance by enforcing block-triangular maps, enabling KV cache sharing and full reuse of intermediate computations.
We apply \method{} at the 1B scale, demonstrating practical speedups and establishing prefix invariance as a useful foundation for adaptive computation. 
\end{abstract}

\vspace{-1.5em}
\section{Introduction}
\vspace{-1.em}

Efficient text generation is a key challenge in deploying Large Language Models (LLMs). These models typically operate \emph{autoregressively}, processing the prompt in parallel during \emph{prefill} and storing its representations in a key-value (KV) cache. During \emph{decode}, the latest token passes through the network, attending to cached representations to predict the next token and extending the cache. Although caching avoids recomputing past representations, generation remains costly because each decoding step requires a full model pass and depends on the token produced by the previous step.

Efforts to reduce this cost build on the observation that many predictions can be handled by smaller, less expensive models. \textit{Routing} exploits this variability by selecting a model for an entire request \citep{ong2025routellm} or individual tokens \citep{zheng2024citer}, while \textit{cascading inference} starts with a smaller model and invokes a larger one when its estimated prediction quality is not satisfactory \citep{narasimhan2025faster}. \textit{Speculative decoding} uses a smaller model to draft tokens that a larger model verifies in parallel \citep{leviathan2023faster}.
However, efficiency depends not only on which model evaluates each token, but also on whether computation can be reused across models. Without such reuse, invoking a larger model may require rebuilding cached representations of the accumulated context, even when only the next prediction requires additional capacity.

This motivates \emph{progressive inference}, in which representations are incrementally refined while retaining prior computation \citep{zhang2018progressiveneuralnetworksimage}, enabling reuse both within a prediction and across the context.
\begin{figure}[t]
\centering
\includegraphics[width=0.95\linewidth]{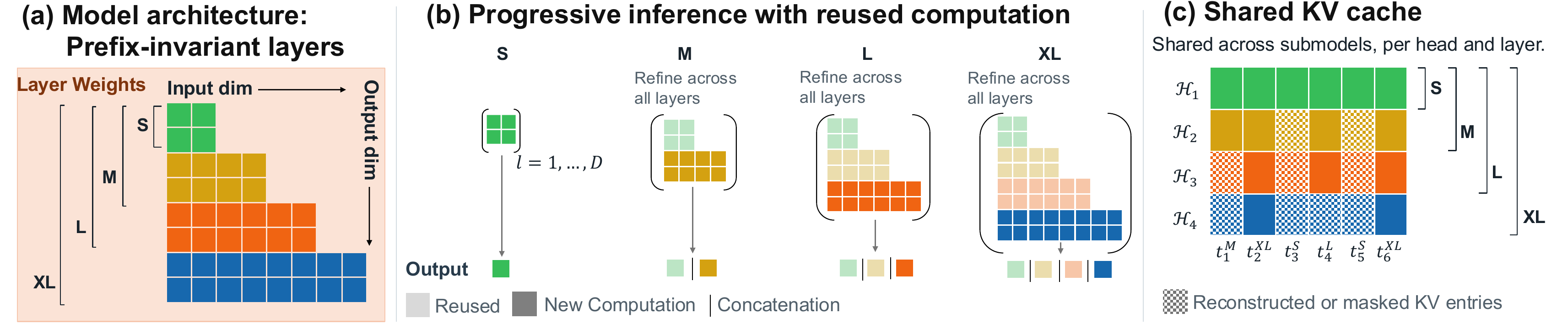}
\caption{\textbf{Progressive inference with \method{}.}
\textbf{(a)} Block lower-triangular weights make every layer prefix invariant: \textbf{(b)} escalating from a smaller submodel to a larger one augments existing representations by activating only the next refinement block. \textbf{(c)} Shared KV caches can be progressively augmented: lacking KV entries for tokens processed by smaller submodel (superscripts) can be exactly computed (checkerboard), or approximated by the other cache policies of \cref{sec:method:cache}.}
\label{fig:slim_to_triod}
\end{figure}
\textit{Early exiting} provides an established route to progressive inference, allowing predictions at intermediate layers while retaining their computation if deeper execution is needed \citep{schuster2022confident,elhoushi2024layerskip}. While these approaches adapt computation along depth, deeper, narrower Transformers can offer favorable performance–cost trade-offs \citep{tay2022scale}. This motivates a complementary question: \textit{Can progressive inference operate along width while retaining the network's full depth?}

Elastic models already adapt computation through nested channel widths \citep{yu2019slimmable}, feed-forward widths \citep{devvrit2024matformer}, or low-rank components \citep{zaccone2026flexrank}. However, parameter nesting does not ensure representation nesting. In slimmable networks, additional channels alter existing activations and thus subsequent layer inputs \citep{yu2019universally}. MatFormer shares cached context between a narrower drafter and a wider verifier, but verification still recomputes the draft tokens' representations and replaces their cached states \citep{devvrit2024matformer}. Progressive inference requires preserving existing representations while computing only the missing components.

We identify this property as \emph{prefix invariance}: smaller submodels' representations are exact prefixes of larger ones at every layer. Width expansion thus refines a coherent representation hierarchy, preserving activations and KV-cache blocks while computing missing components. Unlike Matryoshka embeddings, which make final-vector prefixes independently useful \citep{kusupati2022matryoshka}, prefix invariance requires agreement across separately executed submodels at every layer.

\method{} enforces prefix invariance by construction through block-triangular maps, making the first $d$ output coordinates depend only on the first $d$ input coordinates at every layer. This structure, which we prove necessary and sufficient for prefix invariance, allows width expansion to add computation without revising existing representations (see \cref{fig:slim_to_triod}c).
In contrast, standard slimmable networks \citep{yu2019slimmable} fail to attain prefix invariance even when trained with explicit consistency losses, as wider activations can perturb narrower representations (see \cref{fig:negative}).

Concretely, we make the following contributions:
\begin{itemize}[leftmargin=*, topsep=0pt, itemsep=-1ex, partopsep=1ex, parsep=1ex]
\item We identify \emph{prefix invariance} as the property that enables progressive width expansion through the exact reuse of activations and KV-cache blocks.
\item We introduce \method{}, an architectural design that enforces prefix invariance via block-triangular maps, and connect it to the truncated SVD \citep{rippel2014nested}.
\item We empirically show that explicit consistency loss fails to enforce prefix invariance in standard slimmable networks \citep{yu2019slimmable}, whereas \method{} achieves inference speedups and favorable cost-quality trade-offs through the reuse of hidden states and KV-cache blocks.
\end{itemize}

\vspace{-1.em}
\section{Preliminaries}
\label{sec:preliminaries}
\vspace{-1.em}

This section introduces reusable representations, showing why simply slimming a layer \citep{yu2019slimmable} does not translate nested parameters to shared representations.
The required constraint is prefix invariance, enforced in neural networks by triangular dependencies in the functional map between input and output: the property is preserved under composition, so a model made of such weights is end-to-end prefix-invariant.
This property is crucial for progressive inference: we show a realistic experiment where Slimmable Networks fail at achieving it, even augmenting the objective function with prefix-matching losses, while it holds for \method{} by construction. 

\textbf{From shared parameters to reusable representations.}
Consider an affine layer $y=Wx+b$, with input and output features partitioned into $K$ ordered blocks. A smaller submodel uses the first $s$ blocks. In the full layer, the corresponding outputs receive contributions from both the retained input blocks and those available only at larger widths:
\begin{equation}
    y_{\le s}
    = W_{\le s,\le s}x_{\le s}
    + W_{\le s,>s}x_{>s}
    + b_{\le s}.
    \label{eq:dense-prefix-interference}
\end{equation}
The cross-block term generally changes existing outputs when additional inputs become active.
For a fixed ordered partition, let $P_s$ extract the first $s$ blocks and let $M_s=P_s^\top P_s$ retain those blocks while setting the remaining coordinates to zero. 

\begin{definition}[Prefix invariance]
\label{def:pi}
A map $F:\mathbb{R}^{D}\to\mathbb{R}^{E}$ is \emph{prefix invariant} with respect to its input and output partitions if, for every input $x$ and budget $s\in\{1,\ldots,K\}$,
\begin{equation}
\textstyle
    P_s^{\mathrm{out}}F(x)
    = P_s^{\mathrm{out}}F(M_s^{\mathrm{in}}x).
    \label{eq:pi}
\end{equation}
\end{definition}
The corresponding width-$s$ operation is the restriction
$F^{(s)}(u)=P_s^{\mathrm{out}}F((P_s^{\mathrm{in}})^\top u)$.
Prefix invariance ensures that evaluating this restricted operation produces exactly the matching prefix of the full operation: it imposes a simple restriction on how information can flow between blocks.

\begin{proposition}[Prefix invariance requires triangular maps]
\label{prop:prefix-characterization}
A map $F=(F_1,\ldots, F_K)$ is prefix invariant if and only if each output block can be written as
$
    F_r(x)=f_r(x_1,\ldots,x_r),
    \; r=1,\ldots,K.
$
\end{proposition}
The triangular constraint concerns the complete layer function, relative to the chosen ordering and supported width boundaries.
\Cref{sec:method:blocks} describes how each operation in a transformer block is modified in \method{} to make the overall function prefix-invariant. 

\textbf{Composition.}
Prefix invariance is preserved under composition with aligned intermediate partitions, aligned residual addition, and coordinate-wise nonlinearities. Consider a network of these operations in which each submodel evaluates its corresponding part, starting from a common input embedding. 
Let $h^{\ell,(s)}(u)$ denote its hidden representation at depth $\ell$, evaluated at budget $s$ on input $u$. Applying prefix invariance successively through the network gives
$
    h^{\ell,(s)}(u)
    = h_{\le s}^{\ell,(t)}(u),
    \; s\le t, 
$ for all layers $\ell$, i.e., all submodel partitions can be extracted.

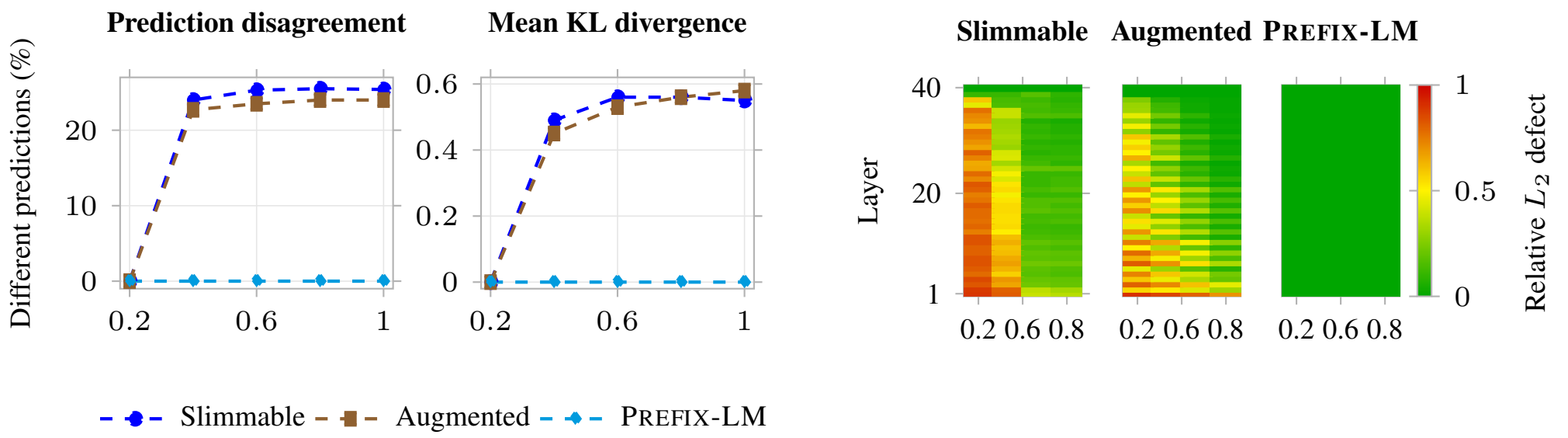
\begin{figure}[t]
\centering
\vspace{-0.8em}
\begin{minipage}[t]{0.49\linewidth}
\vspace{0pt}\centering
\begin{tikzpicture}
\begin{groupplot}[
  paper pair,width=0.34\linewidth,
  group style={group size=2 by 1,horizontal sep=\PlotGap},
  xmin=0.17,xmax=1.03,xtick={0.2,0.6,1.0},
  every axis plot/.append style={dashed},
  legend columns=3
]
\nextgroupplot[legend to name=slimlegend,title={Prediction disagreement},ylabel={Different predictions (\%)},
  ymin=-1,ymax=27,ytick={0,10,20}]
\addplot[blue,mark=*] coordinates
 {(0.2,0) (0.4,24.0) (0.6,25.3) (0.8,25.5) (1.0,25.4)};
\addlegendentry{Slimmable}
\addplot[brown!75!black,mark=square*] coordinates
 {(0.2,0) (0.4,22.7) (0.6,23.5) (0.8,24.0) (1.0,24.0)};
\addlegendentry{Augmented}
\addplot[cyan!85!blue,mark=diamond*] coordinates
 {(0.2,0) (0.4,0) (0.6,0) (0.8,0) (1.0,0)};
\addlegendentry{\method{}}
\nextgroupplot[title={Mean KL divergence},
  ymin=-0.02,ymax=0.62,ytick={0,0.2,0.4,0.6}]
\addplot[blue,mark=*] coordinates
 {(0.2,0) (0.4,0.49) (0.6,0.56) (0.8,0.56) (1.0,0.55)};
\addplot[brown!75!black,mark=square*] coordinates
 {(0.2,0) (0.4,0.45) (0.6,0.53) (0.8,0.56) (1.0,0.58)};
\addplot[cyan!85!blue,mark=diamond*] coordinates
 {(0.2,0) (0.4,0) (0.6,0) (0.8,0) (1.0,0)};
\end{groupplot}
\path (group c1r1.south) -- (group c2r1.south)
  node[midway,anchor=north,inner sep=0pt,yshift=-\PlotLegendGap]
  {\pgfplotslegendfromname{slimlegend}};
\end{tikzpicture}
\end{minipage}%
\quad
\begin{minipage}[t]{0.45\linewidth}
\vspace{0pt}\centering
\begin{tikzpicture}
\begin{groupplot}[
  group style={group size=3 by 1,horizontal sep=0.35cm},
  paper axis,width=0.16\linewidth,grid=none,enlargelimits=false,
  title style={text height=1.5ex,text depth=0.3ex},
  xmin=0,xmax=4,ymin=0,ymax=40,
  xtick={0.5,2,3.5},xticklabels={0.2,0.6,0.8},
  ytick=\empty,
  colormap={goodbad}{color(0cm)=(green!65!black);color(1cm)=(yellow!90!orange);color(2cm)=(red!80!black)},
  point meta min=0,point meta max=1
]
\nextgroupplot[title={Slimmable},ylabel={Layer},ytick={0.5,19.5,39.5},yticklabels={1,20,40}]
\addplot[matrix plot*,mesh/cols=4,mesh/rows=40,point meta=explicit]
table[meta=z] {x y z
0.5 0.5 0.89
1.5 0.5 0.81
2.5 0.5 0.39
3.5 0.5 0.36
0.5 1.5 0.88
1.5 1.5 0.80
2.5 1.5 0.37
3.5 1.5 0.32
0.5 2.5 0.82
1.5 2.5 0.66
2.5 2.5 0.22
3.5 2.5 0.20
0.5 3.5 0.85
1.5 3.5 0.69
2.5 3.5 0.20
3.5 3.5 0.19
0.5 4.5 0.78
1.5 4.5 0.57
2.5 4.5 0.17
3.5 4.5 0.15
0.5 5.5 0.81
1.5 5.5 0.63
2.5 5.5 0.18
3.5 5.5 0.16
0.5 6.5 0.83
1.5 6.5 0.67
2.5 6.5 0.22
3.5 6.5 0.21
0.5 7.5 0.83
1.5 7.5 0.65
2.5 7.5 0.18
3.5 7.5 0.17
0.5 8.5 0.80
1.5 8.5 0.61
2.5 8.5 0.19
3.5 8.5 0.18
0.5 9.5 0.80
1.5 9.5 0.60
2.5 9.5 0.16
3.5 9.5 0.14
0.5 10.5 0.83
1.5 10.5 0.64
2.5 10.5 0.19
3.5 10.5 0.17
0.5 11.5 0.83
1.5 11.5 0.61
2.5 11.5 0.15
3.5 11.5 0.14
0.5 12.5 0.74
1.5 12.5 0.51
2.5 12.5 0.13
3.5 12.5 0.12
0.5 13.5 0.79
1.5 13.5 0.55
2.5 13.5 0.12
3.5 13.5 0.11
0.5 14.5 0.77
1.5 14.5 0.54
2.5 14.5 0.14
3.5 14.5 0.15
0.5 15.5 0.78
1.5 15.5 0.54
2.5 15.5 0.13
3.5 15.5 0.13
0.5 16.5 0.78
1.5 16.5 0.55
2.5 16.5 0.18
3.5 16.5 0.16
0.5 17.5 0.81
1.5 17.5 0.55
2.5 17.5 0.16
3.5 17.5 0.13
0.5 18.5 0.76
1.5 18.5 0.53
2.5 18.5 0.16
3.5 18.5 0.16
0.5 19.5 0.74
1.5 19.5 0.48
2.5 19.5 0.11
3.5 19.5 0.11
0.5 20.5 0.80
1.5 20.5 0.55
2.5 20.5 0.14
3.5 20.5 0.16
0.5 21.5 0.82
1.5 21.5 0.53
2.5 21.5 0.11
3.5 21.5 0.13
0.5 22.5 0.72
1.5 22.5 0.43
2.5 22.5 0.11
3.5 22.5 0.13
0.5 23.5 0.78
1.5 23.5 0.48
2.5 23.5 0.11
3.5 23.5 0.11
0.5 24.5 0.66
1.5 24.5 0.41
2.5 24.5 0.20
3.5 24.5 0.19
0.5 25.5 0.65
1.5 25.5 0.32
2.5 25.5 0.13
3.5 25.5 0.09
0.5 26.5 0.73
1.5 26.5 0.43
2.5 26.5 0.11
3.5 26.5 0.11
0.5 27.5 0.80
1.5 27.5 0.42
2.5 27.5 0.08
3.5 27.5 0.09
0.5 28.5 0.66
1.5 28.5 0.30
2.5 28.5 0.10
3.5 28.5 0.09
0.5 29.5 0.70
1.5 29.5 0.32
2.5 29.5 0.08
3.5 29.5 0.07
0.5 30.5 0.74
1.5 30.5 0.34
2.5 30.5 0.11
3.5 30.5 0.09
0.5 31.5 0.76
1.5 31.5 0.32
2.5 31.5 0.09
3.5 31.5 0.07
0.5 32.5 0.62
1.5 32.5 0.31
2.5 32.5 0.10
3.5 32.5 0.07
0.5 33.5 0.70
1.5 33.5 0.30
2.5 33.5 0.07
3.5 33.5 0.05
0.5 34.5 0.72
1.5 34.5 0.39
2.5 34.5 0.16
3.5 34.5 0.15
0.5 35.5 0.77
1.5 35.5 0.39
2.5 35.5 0.13
3.5 35.5 0.11
0.5 36.5 0.45
1.5 36.5 0.15
2.5 36.5 0.14
3.5 36.5 0.10
0.5 37.5 0.59
1.5 37.5 0.14
2.5 37.5 0.12
3.5 37.5 0.08
0.5 38.5 0.11
1.5 38.5 0.11
2.5 38.5 0.17
3.5 38.5 0.09
0.5 39.5 0.00
1.5 39.5 0.00
2.5 39.5 0.00
3.5 39.5 0.00
};
\nextgroupplot[title={Augmented}]
\addplot[matrix plot*,mesh/cols=4,mesh/rows=40,point meta=explicit]
table[meta=z] {x y z
0.5 0.5 0.90
1.5 0.5 0.86
2.5 0.5 0.80
3.5 0.5 0.70
0.5 1.5 0.56
1.5 1.5 0.52
2.5 1.5 0.43
3.5 1.5 0.31
0.5 2.5 0.80
1.5 2.5 0.72
2.5 2.5 0.60
3.5 2.5 0.45
0.5 3.5 0.59
1.5 3.5 0.46
2.5 3.5 0.33
3.5 3.5 0.20
0.5 4.5 0.73
1.5 4.5 0.59
2.5 4.5 0.41
3.5 4.5 0.23
0.5 5.5 0.45
1.5 5.5 0.30
2.5 5.5 0.17
3.5 5.5 0.08
0.5 6.5 0.81
1.5 6.5 0.70
2.5 6.5 0.54
3.5 6.5 0.34
0.5 7.5 0.63
1.5 7.5 0.48
2.5 7.5 0.31
3.5 7.5 0.15
0.5 8.5 0.78
1.5 8.5 0.65
2.5 8.5 0.48
3.5 8.5 0.28
0.5 9.5 0.50
1.5 9.5 0.35
2.5 9.5 0.20
3.5 9.5 0.09
0.5 10.5 0.74
1.5 10.5 0.65
2.5 10.5 0.48
3.5 10.5 0.27
0.5 11.5 0.39
1.5 11.5 0.25
2.5 11.5 0.13
3.5 11.5 0.05
0.5 12.5 0.69
1.5 12.5 0.53
2.5 12.5 0.35
3.5 12.5 0.17
0.5 13.5 0.45
1.5 13.5 0.28
2.5 13.5 0.15
3.5 13.5 0.05
0.5 14.5 0.54
1.5 14.5 0.44
2.5 14.5 0.29
3.5 14.5 0.15
0.5 15.5 0.39
1.5 15.5 0.20
2.5 15.5 0.10
3.5 15.5 0.04
0.5 16.5 0.71
1.5 16.5 0.55
2.5 16.5 0.35
3.5 16.5 0.17
0.5 17.5 0.57
1.5 17.5 0.37
2.5 17.5 0.18
3.5 17.5 0.06
0.5 18.5 0.69
1.5 18.5 0.50
2.5 18.5 0.30
3.5 18.5 0.14
0.5 19.5 0.50
1.5 19.5 0.30
2.5 19.5 0.15
3.5 19.5 0.05
0.5 20.5 0.69
1.5 20.5 0.52
2.5 20.5 0.33
3.5 20.5 0.16
0.5 21.5 0.37
1.5 21.5 0.19
2.5 21.5 0.09
3.5 21.5 0.03
0.5 22.5 0.60
1.5 22.5 0.39
2.5 22.5 0.20
3.5 22.5 0.08
0.5 23.5 0.39
1.5 23.5 0.19
2.5 23.5 0.08
3.5 23.5 0.02
0.5 24.5 0.50
1.5 24.5 0.28
2.5 24.5 0.12
3.5 24.5 0.03
0.5 25.5 0.33
1.5 25.5 0.13
2.5 25.5 0.05
3.5 25.5 0.01
0.5 26.5 0.64
1.5 26.5 0.40
2.5 26.5 0.19
3.5 26.5 0.07
0.5 27.5 0.50
1.5 27.5 0.23
2.5 27.5 0.08
3.5 27.5 0.02
0.5 28.5 0.57
1.5 28.5 0.29
2.5 28.5 0.12
3.5 28.5 0.03
0.5 29.5 0.48
1.5 29.5 0.19
2.5 29.5 0.06
3.5 29.5 0.01
0.5 30.5 0.59
1.5 30.5 0.33
2.5 30.5 0.16
3.5 30.5 0.05
0.5 31.5 0.28
1.5 31.5 0.10
2.5 31.5 0.04
3.5 31.5 0.01
0.5 32.5 0.48
1.5 32.5 0.23
2.5 32.5 0.08
3.5 32.5 0.02
0.5 33.5 0.30
1.5 33.5 0.10
2.5 33.5 0.03
3.5 33.5 0.01
0.5 34.5 0.45
1.5 34.5 0.23
2.5 34.5 0.09
3.5 34.5 0.02
0.5 35.5 0.31
1.5 35.5 0.13
2.5 35.5 0.03
3.5 35.5 0.01
0.5 36.5 0.30
1.5 36.5 0.12
2.5 36.5 0.03
3.5 36.5 0.01
0.5 37.5 0.25
1.5 37.5 0.07
2.5 37.5 0.02
3.5 37.5 0.01
0.5 38.5 0.00
1.5 38.5 0.00
2.5 38.5 0.00
3.5 38.5 0.00
0.5 39.5 0.00
1.5 39.5 0.00
2.5 39.5 0.00
3.5 39.5 0.00
};
\nextgroupplot[title={\method{}},colorbar,
 colorbar style={at={(1.15,0.5)},anchor=west,width=0.12cm,height=\PlotHeight,
 ytick={0,0.5,1},ylabel={Relative $L_2$ defect},
 y label style={font=\PlotFont,xshift=-3pt},yticklabel style={font=\PlotFont}}]
\addplot[matrix plot*,mesh/cols=4,mesh/rows=40,point meta=explicit]
table[meta=z] {x y z
0.5 0.5 0.00
1.5 0.5 0.00
2.5 0.5 0.00
3.5 0.5 0.00
0.5 1.5 0.00
1.5 1.5 0.00
2.5 1.5 0.00
3.5 1.5 0.00
0.5 2.5 0.00
1.5 2.5 0.00
2.5 2.5 0.00
3.5 2.5 0.00
0.5 3.5 0.00
1.5 3.5 0.00
2.5 3.5 0.00
3.5 3.5 0.00
0.5 4.5 0.00
1.5 4.5 0.00
2.5 4.5 0.00
3.5 4.5 0.00
0.5 5.5 0.00
1.5 5.5 0.00
2.5 5.5 0.00
3.5 5.5 0.00
0.5 6.5 0.00
1.5 6.5 0.00
2.5 6.5 0.00
3.5 6.5 0.00
0.5 7.5 0.00
1.5 7.5 0.00
2.5 7.5 0.00
3.5 7.5 0.00
0.5 8.5 0.00
1.5 8.5 0.00
2.5 8.5 0.00
3.5 8.5 0.00
0.5 9.5 0.00
1.5 9.5 0.00
2.5 9.5 0.00
3.5 9.5 0.00
0.5 10.5 0.00
1.5 10.5 0.00
2.5 10.5 0.00
3.5 10.5 0.00
0.5 11.5 0.00
1.5 11.5 0.00
2.5 11.5 0.00
3.5 11.5 0.00
0.5 12.5 0.00
1.5 12.5 0.00
2.5 12.5 0.00
3.5 12.5 0.00
0.5 13.5 0.00
1.5 13.5 0.00
2.5 13.5 0.00
3.5 13.5 0.00
0.5 14.5 0.00
1.5 14.5 0.00
2.5 14.5 0.00
3.5 14.5 0.00
0.5 15.5 0.00
1.5 15.5 0.00
2.5 15.5 0.00
3.5 15.5 0.00
0.5 16.5 0.00
1.5 16.5 0.00
2.5 16.5 0.00
3.5 16.5 0.00
0.5 17.5 0.00
1.5 17.5 0.00
2.5 17.5 0.00
3.5 17.5 0.00
0.5 18.5 0.00
1.5 18.5 0.00
2.5 18.5 0.00
3.5 18.5 0.00
0.5 19.5 0.00
1.5 19.5 0.00
2.5 19.5 0.00
3.5 19.5 0.00
0.5 20.5 0.00
1.5 20.5 0.00
2.5 20.5 0.00
3.5 20.5 0.00
0.5 21.5 0.00
1.5 21.5 0.00
2.5 21.5 0.00
3.5 21.5 0.00
0.5 22.5 0.00
1.5 22.5 0.00
2.5 22.5 0.00
3.5 22.5 0.00
0.5 23.5 0.00
1.5 23.5 0.00
2.5 23.5 0.00
3.5 23.5 0.00
0.5 24.5 0.00
1.5 24.5 0.00
2.5 24.5 0.00
3.5 24.5 0.00
0.5 25.5 0.00
1.5 25.5 0.00
2.5 25.5 0.00
3.5 25.5 0.00
0.5 26.5 0.00
1.5 26.5 0.00
2.5 26.5 0.00
3.5 26.5 0.00
0.5 27.5 0.00
1.5 27.5 0.00
2.5 27.5 0.00
3.5 27.5 0.00
0.5 28.5 0.00
1.5 28.5 0.00
2.5 28.5 0.00
3.5 28.5 0.00
0.5 29.5 0.00
1.5 29.5 0.00
2.5 29.5 0.00
3.5 29.5 0.00
0.5 30.5 0.00
1.5 30.5 0.00
2.5 30.5 0.00
3.5 30.5 0.00
0.5 31.5 0.00
1.5 31.5 0.00
2.5 31.5 0.00
3.5 31.5 0.00
0.5 32.5 0.00
1.5 32.5 0.00
2.5 32.5 0.00
3.5 32.5 0.00
0.5 33.5 0.00
1.5 33.5 0.00
2.5 33.5 0.00
3.5 33.5 0.00
0.5 34.5 0.00
1.5 34.5 0.00
2.5 34.5 0.00
3.5 34.5 0.00
0.5 35.5 0.00
1.5 35.5 0.00
2.5 35.5 0.00
3.5 35.5 0.00
0.5 36.5 0.00
1.5 36.5 0.00
2.5 36.5 0.00
3.5 36.5 0.00
0.5 37.5 0.00
1.5 37.5 0.00
2.5 37.5 0.00
3.5 37.5 0.00
0.5 38.5 0.00
1.5 38.5 0.00
2.5 38.5 0.00
3.5 38.5 0.00
0.5 39.5 0.00
1.5 39.5 0.00
2.5 39.5 0.00
3.5 39.5 0.00
};
\end{groupplot}
\end{tikzpicture}
\end{minipage}
\vspace{-0.8em}
{\small\caption{\textbf{Slimmable networks are not progressive.} Prediction disagreement (left) and layerwise prefix discrepancy (right) at relative width $p$ for slimmable, an augmented recipe, and \method{}, on CIFAR-100 and ResNet-20.
Despite augmenting the objective with prefix-constraining losses, slimmable networks do not align intermediate representations, accumulating errors over layers
\label{fig:negative}}}
\vspace{-1.2em}
\end{figure}

\textbf{Conventional slimmable networks.}
Standard slimmable networks~\citep{yu2019slimmable} are not prefix-invariant because they retain the cross-block dependencies illustrated above. While their training procedures target useful predictions at multiple widths, they do not impose the prefix identity in ~\cref{def:pi}. 
To further validate that prefix-invariance cannot be imposed onto slimmable networks by properly adjusting the training objective, we propose a realistic experiment in which the standard cross-entropy loss is augmented with prefix-consistency penalties. Concretely, we impose an $\ell_2$ penalty between the shared-prefixes of output features of each submodel, across all the layers. 
This is exactly what block triangular weights in \method{} enforce by construction.
Results in \cref{fig:negative}
show that both the standard recipe and the augmented objective function lead to non-negligible prediction disagreement between a progressive-style forward and the actual full recomputation without reuse, corroborating the theoretical result in \cref{prop:prefix-characterization}.

\vspace{-1.em}
\section{\method{}}
\label{sec:method}
\vspace{-1.em}
Here we describe the components of \method{}, depicted in~\cref{fig:slim_to_triod}: (i) the operations of a prefix-invariant transformer block, (ii) the optimized training procedure, matching the FLOPs of a regular one, (iii) efficient KV-cache reuse and approximate alternatives to backfilling for per-token progressive inference. Finally, we discuss the relation to truncated SVD. Proofs are deferred to \cref{sec:appendix:proofs}.

\subsection{Prefix-invariant Transformer blocks}
\label{sec:method:blocks}

We use $K$ ordered refinement blocks throughout the hidden layers. Block sizes may differ between the residual stream and the feed-forward intermediate representation, but their ordering is aligned: Model $s$ retains blocks $1{:}s$ at every layer. The input embedding remains dense because every budget receives the same raw input. Hidden transformations preserve prefix invariance, and a shared dense classifier accumulates the contributions of the final representation blocks.

\textbf{Linear maps and feed-forward layers.}
Following \cref{prop:prefix-characterization}, let $W_{rq}$ map input block $q$ to output block $r$. 
For a hidden affine map $F(x)=Wx+b$, it is sufficient to set $W_{rq}=0$ for $q>r$, where $W_{rq}$ maps input block $q$ to output block $r$ or, equivalently, $y_r=\sum_{q=1}^{r}W_{rq}x_q+b_r$. We apply this ordering to both feed-forward projections and, for gated feed-forward layers, to both the feature and gate branches. Their element-wise product preserves the block structure.

\textbf{Hierarchical normalization.}
Standard normalization operators generally violate prefix invariance, since all the feature dimension blocks contribute to the calculation of the statistics. For this reason, for each block $r$ the operation should be constrained to the cumulative feature dimension \mbox{$D_r=\sum_{q\le r}d_q$}. Then, for each token, we define the normalization statistics of block $r$ as:
\begin{equation}
\small %
\textstyle
\mu_{:r}(x)=\frac{1}{D_r}\sum_{i\le D_r}x_i,
\qquad
\sigma_{:r}^{2}(x)=\frac{1}{D_r}\sum_{i\le D_r}(x_i-\mu_{:r}(x))^2.
\label{eq:hln-moments}
\end{equation}

\textbf{Nested multi-head attention.}
Multi-Head Attention is modified by partitioning heads into groups $\mathcal H_1,\ldots,\mathcal H_K$; model $s$ activates $\cup_{q\le s}\mathcal H_q$, with head dimensions fixed across submodels. Let $X$ contain the normalized embeddings as rows, and write $X_{\le q}=X(P_q^{\mathrm{in}})^\top$. For a head $h\in\mathcal H_q$, its query, key, and value projections read only $X_{\le q}$. With a common causal additive mask $C$, its output and the subsequent block projection are
\begin{equation*}
\small
\textstyle
O_h(X)=\operatorname{softmax}\!\left(
\frac{Q_h(X_{\le q})K_h(X_{\le q})^\top}{\sqrt{d_h}}+C
\right)V_h(X_{\le q}),\;
a_r(X)=\sum_{q\le r}
\operatorname{concat}_{h\in\mathcal H_q}\!\bigl[O_h(X)\bigr]
(W^O_{rq})^\top,
\end{equation*}
Each earlier head retains its queries, keys, values, attention weights, and output as later groups are activated. 
The triangular output projection prevents later head groups from modifying earlier residual-stream blocks. For grouped- or multi-query attention~\citep{ainslie2023gqa,shazeer2019fast}, the query heads sharing a key/value head must belong to the same group, so that the shared head is computed by the smallest budget that uses it and remains valid for all larger ones.

\textbf{Cumulative logits.}
Prefix invariance of features at the last transformer block allows a key property, making progressive inference efficient also for the final vocabulary projection: the output logits can be computed incrementally for each submodel, as activating block $s+1$ adds a logit refinement without recomputing the earlier contributions. More formally, let $h^L=(h_1^L,\ldots,h_K^L)$ denote the final representation and partition the shared (dense) classifier as $A=[A_1\;\cdots\;A_K]$. Its logits are
\begin{equation}
    \textstyle
    z_s(x)=b+\sum_{r=1}^{s}A_rh_r^L(x),
    \qquad
    z_{s+1}(x)=z_s(x)+A_{s+1}h_{s+1}^L(x).
\label{eq:cumulative-logits}
\end{equation}
This is relevant because it avoids repeating the cost of the final head multiplication when escalating to a bigger submodel, a typical overhead of non-exited heads in early-exiting. 

\textbf{Prefix-invariant Mamba blocks}
For the Nemotron-style
Mamba-2 mixer, partition expanded channels into groups $\mathcal H_1,\ldots,\mathcal H_K$ of complete recurrent heads. A head $h\in\mathcal H_r$ reads only the residual-stream prefix through block $r$ for its input, gate, and step size. Its full-dimensional $B_t,C_t$ vectors are shared across heads and depend only on the smallest prefix. With causal depthwise-convolved inputs $\bar x_{h,t},\bar B_t,\bar C_t$ and step size $\Delta_{h,t}$, each head updates its state and output as
\begin{equation}
\begin{aligned}
\textstyle
S_{h,t}&=\exp(\Delta_{h,t}A_h)S_{h,t-1}
       +\Delta_{h,t}\,\bar x_{h,t}\bar B_t^{\top},\qquad
o_{h,t}&=S_{h,t}\bar C_t+D_h\bar x_{h,t}.
\end{aligned}
\label{eq:triod-mamba-scan}
\end{equation}
The gate, hierarchical RMS normalization, and output projection preserve the same prefix ordering. Existing heads' convolution and recurrent states are therefore width-invariant for a common token history and mask. 
Escalation still requires backfilling the missing features and states of newly activated blocks through the earlier layers. Full formulas and the argument are in \cref{app:mamba-block}.

\textbf{Architectural guarantees.} The above constructions lead to prefix-invariant transformers.
\begin{theorem}[Component invariance]
\label{thm:method-components}
The affine maps, hierarchical normalization, and nested attention are all prefix invariant.
\end{theorem}

\begin{theorem}[Composability]
\label{thm:method-composition}
Prefix-invariant maps compose when their intermediate partitions are aligned. The class is also closed under aligned addition, coordinate-wise nonlinearities, and concatenation within corresponding blocks.
\end{theorem}

\begin{corollary}[Single-pass exactness]
\label{cor:method-singlepass}
At every layer of the construction above, the first $s$ blocks of a full-width backbone produce outputs equal to those produced by evaluating submodel $s$ alone, on the same input and with the same attention masks. The cumulative classifier in \cref{eq:cumulative-logits} consequently recovers every submodel's logits from that pass.
\end{corollary}

\subsection{Training \method{}}
\label{sec:training}
\input{figures/hkd_vs_kd}
Prefix invariance also shapes how \method{} is trained. Because every submodel's representation is a prefix of the full model's, a single forward pass exposes all $K$ predictions, and any per-budget objective can be optimized over one shared computation graph. This raises two questions: (i) how to supervise the smaller budgets, and (ii) how to weight the parameters that participate in only a subset of them. We address the first with a \emph{hierarchical self-distillation objective} that trains each submodel from its immediate successor, and the second with a \emph{membership-normalized gradient} that equalizes the participation scaling introduced by joint training. Both use quantities already available from the single pass, so their overhead is confined to the readout.

\textbf{Training all models in a single pass.}
By \cref{cor:method-singlepass}, one full-width forward pass yields the logits $z_1,\ldots,z_K$ of every submodel, so all per-budget losses can be applied to a single computation graph and backpropagated together. This contrasts with the training of slimmable networks~\citep{yu2019universally}, which introduces submodel sampling at each step.
The classifier adds no overhead either, as the cumulative readout in \cref{eq:cumulative-logits} evaluates each product $A_rh_r^L$ once, so its total cost equals that of a single full-width classifier. The only additional work relative to a vanilla transformer lies in the readout: accumulating prefix logits, normalizing each $z_s$ over the vocabulary, and evaluating $K$ losses instead of one. 

\textbf{Hierarchical Self-Distillation (\hsd{}).}
A common way to supervise the smaller budgets of a many-in-one model is self-distillation~\citep{yu2019universally,zaccone2026flexrank}. Beyond transferring predictive knowledge, distillation also serves as a form of partial variance reduction: \citet{safaryan2023knowledge} show that the variance of the student's stochastic gradient scales with the discrepancy between student and teacher predictions. This suggests supervising each submodel from the closest available teacher rather than the full model. In \method{}, the natural choice is the next larger budget, which differs from the student by exactly one refinement block and therefore provides a target that is both stronger than the student and closer to it than the full model. We call this \emph{hierarchical self-distillation} (HSD). For $K\ge2$, let $p_s=\operatorname{softmax}(z_s)$ and let $\operatorname{sg}$ stop teacher gradients:
\begin{equation}
\textstyle
\loss_{\mathrm{HSD}}=
\frac{1}{K-1}\sum_{s=1}^{K-1}
D_{\mathrm{KL}}\!\left(\operatorname{sg}[p_{s+1}]\,\middle\|\,p_s\right),
\qquad
\loss=\frac{1}{K}\left(\operatorname{CE}(z_K,y)+(K-1)\loss_{\mathrm{HSD}}\right).
\label{eq:hkd}
\end{equation}
Only the full model receives label supervision; all other budgets follow the detached distribution of their successor, and all teachers come from the same forward pass, so HSD requires no additional teacher model or pass. HSD is also cheap to backpropagate (see \cref{app:telescoping}). We visualize its improved gradient variance and convergence in \Cref{fig:hkd_vs_kd} on nested low-rank regression, which is equivalent to solving a truncated SVD along each width (\cref{app:hsd_vs_sd}).

\textbf{Membership-Normalized Gradients (\mng{}).}
Joint training introduces a structural imbalance, as parameters first activated at budget $r$ receive gradients only from the losses of budgets $s\ge r$.
Writing $\ell_s$ for budget $s$ in \cref{eq:hkd}, the overall gradient with respect to such parameters $\theta_r$ is
\begin{equation}
\textstyle
\nabla_{\theta_r}\loss
=\frac{1}{K}\sum_{s=r}^{K}\nabla_{\theta_r}\ell_s
=c_r\cdot\frac{1}{K-r+1}\sum_{s=r}^{K}\nabla_{\theta_r}\ell_s,
\qquad c_r=\frac{K-r+1}{K},
\label{eq:dilution}
\end{equation}
that is, $c_r$ times the mean gradient over the objectives in which $\theta_r$ participates.
Late blocks are therefore updated with systematically smaller steps, down to a factor $1/K$ for the last one, for no reason other than submodel selection.
\mng{} removes this factor by rescaling each block's gradient by $1/c_r$, thereby acting as a blockwise preconditioner to equalize participation scales.
\begin{revised}
Whether this rescaling changes the update depends on the optimizer.
AdamW~\citep{loshchilov2019decoupled} normalizes each coordinate by the running magnitude of its own gradient, so a constant per-block factor cancels up to the effect of $\epsilon$; \mng{} therefore has little effect in our language-model experiments, which use AdamW.
It matters for optimizers without this invariance.
With SGD, \mng{} accelerates convergence, increasingly so at larger widths (\cref{sec:appendix:mng_synthetic}).
Muon~\citep{jordan2024muon} orthogonalizes the update of each weight matrix as a whole: it is invariant to a uniform scale of the matrix, but not to the different factors $1/c_r$ that \mng{} applies to the blocks within one matrix, so we expect \mng{} to be relevant there as well.
\end{revised}

\subsection{Autoregressive KV-cache reuse}
\label{sec:method:cache}
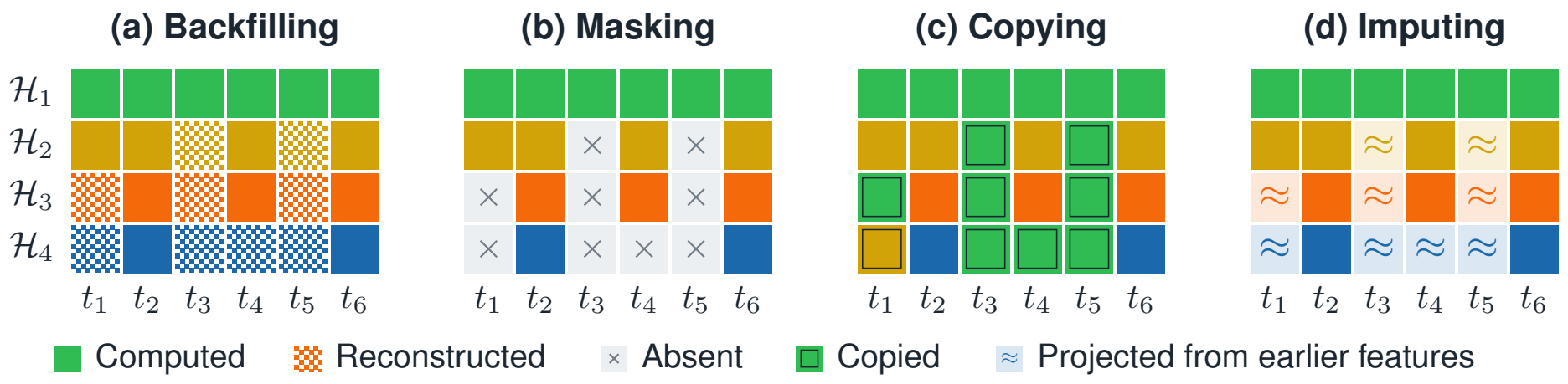
\begin{figure}[t]
\centering
\begingroup

\providecommand{\KVFigureScale}{0.9}

\definecolor{KVHead1}{HTML}{30BB53}
\definecolor{KVHead2}{HTML}{D2A209}
\definecolor{KVHead3}{HTML}{F46A0A}
\definecolor{KVHead4}{HTML}{1C68AC}
\definecolor{KVInk}{HTML}{1B2630}
\definecolor{KVAbsent}{HTML}{ECEFF1}
\definecolor{KVCross}{HTML}{727D87}

\newcommand{\KVReconstructed}[2]{%
  \begin{scope}
    \clip (0,0) rectangle (#2,#2);
    \fill[white] (0,0) rectangle (#2,#2);
    \pgfmathtruncatemacro{\KVLast}{ceil(#2/4)-1}
    \foreach \KVa in {0,...,\KVLast}{%
      \foreach \KVb in {0,...,\KVLast}{%
        \ifodd\numexpr\KVa+\KVb\relax\else
          \fill[#1] ({4*\KVa},{4*\KVb}) rectangle ++(4,4);
        \fi
      }%
    }%
  \end{scope}%
}
\providecommand{\KVProjected}[2]{%
  \fill[#1!16!white] (0,0) rectangle (#2,#2);
  \node[text=#1,scale={.085*#2},font=\fontsize{10}{12}\selectfont]
    at ({.5*#2},{.5*#2}) {$\approx$};%
}
\newcommand{\KVCopied}[2]{%
  \fill[#1] (0,0) rectangle (#2,#2);
  \draw[KVInk,line width=1.1pt] (4,4) rectangle ({#2-4},{#2-4});%
}
\newcommand{\KVMissing}[1]{%
  \fill[KVAbsent] (0,0) rectangle (#1,#1);
  \draw[KVCross,line width=1.5pt]
    ({.33*#1},{.33*#1}) -- ({.67*#1},{.67*#1})
    ({.33*#1},{.67*#1}) -- ({.67*#1},{.33*#1});%
}

\noindent\resizebox{\KVFigureScale\linewidth}{!}{%
\begin{tikzpicture}[
  x=1pt,y=-1pt,
  every node/.style={inner sep=0,outer sep=0,text=KVInk,
    font=\sffamily\fontsize{9}{10}\selectfont},
  kv title/.style={anchor=base,scale=2.9,
    font=\sffamily\bfseries\fontsize{10}{12}\selectfont},
  kv axis/.style={scale=2.9,font=\fontsize{10}{12}\selectfont},
  kv legend/.style={anchor=base west,scale=3}
]
\path[use as bounding box] (0,0) rectangle (1320,326);

\foreach \KVPolicy/\KVTitle in {
  0/{(a) Backfilling},1/{(b) Masking},
  2/{(c) Copying},3/{(d) Imputing}}{%
  \begin{scope}[shift={({16+326*\KVPolicy},0)}]
    \node[kv title] at (169.5,34) {\KVTitle};
    \foreach \KVHead in {1,...,4}{%
      \edef\KVColor{KVHead\KVHead}
      \ifnum\KVPolicy=0
        \node[kv axis,anchor=base east]
          at (28,{86+43*(\KVHead-1)}) {$\mathcal{H}_{\KVHead}$};
      \fi
      \foreach \KVActive [count=\KVToken from 1] in {2,4,1,3,1,4}{%
        \begin{scope}[shift={({42+43*(\KVToken-1)},{60+43*(\KVHead-1)})}]
          \ifnum\KVHead>\KVActive
            \ifcase\KVPolicy
              \KVReconstructed{\KVColor}{40}%
            \or
              \KVMissing{40}%
            \or
              \pgfmathtruncatemacro{\KVSource}{mod(\KVHead-1,\KVActive)+1}
              \edef\KVSourceColor{KVHead\KVSource}
              \KVCopied{\KVSourceColor}{40}%
            \or
              \KVProjected{\KVColor}{40}%
            \fi
          \else
            \fill[\KVColor] (0,0) rectangle (40,40);
          \fi
        \end{scope}%
      }%
    }%
    \foreach \KVToken in {1,...,6}{%
      \node[kv axis,anchor=base]
        at ({62+43*(\KVToken-1)},258) {$t_{\KVToken}$};
    }%
  \end{scope}%
}

\fill[KVHead1] (44,289) rectangle ++(22,22);
\node[kv legend] at (78,307) {Computed};
\begin{scope}[shift={(243,289)}]\KVReconstructed{KVHead3}{22}\end{scope}
\node[kv legend] at (277,307) {Reconstructed};
\begin{scope}[shift={(497,289)}]\KVMissing{22}\end{scope}
\node[kv legend] at (531,307) {Absent};
\begin{scope}[shift={(659,289)}]\KVCopied{KVHead1}{22}\end{scope}
\node[kv legend] at (693,307) {Copied};
\begin{scope}[shift={(825,289)}]\KVProjected{KVHead4}{22}\end{scope}
\node[kv legend] at (859,307) {Projected from earlier features};
\end{tikzpicture}%
}%
\endgroup%
\vspace{-1.em}
{\small
\caption{\textbf{KV cache strategies during progressive decoding visualized.} Rows denote head groups $\mathcal{H}_r$, and solid cells were computed during the original exits at the time of decoding each token. \textbf{(a)} Backfilling reconstructs missing historical KV states. \textbf{(b)} Masking performs sparse attention over absent heads. \textbf{(c)} Copying reads from KV states that belong to earlier blocks (cycling which block) if the entry is otherwise missing. \textbf{(d)} Imputing obtains KV states by applying each missing head's KV projections to the available features at the original exit.
}
\label{fig:kv-cache-policies}
}
\end{figure}

KV cache reuse in autoregressive language modeling follows as a straightforward consequence of prefix invariance. 
Switching budgets during decoding, therefore, never invalidates existing cache entries.
Methods without prefix invariance offer no such guarantee: their shared heads produce different keys and values at different widths, so a width change forces the receiving model to rebuild or approximate the context~\citep{devvrit2024matformer,liu2026droidspeak}.

However, in routing or model cascades, given the current token processed with budget $t>s$ it is possible that a previous token was processed at budget $s$, missing cache entries for the heads in $\mathcal H_{s+1},\ldots,\mathcal H_t$ (\cref{fig:slim_to_triod}c).
\method{} supports a family of policies for these entries (\cref{fig:kv-cache-policies}). \emph{Backfilling} is exact: since blocks $1{:}s$ of the past tokens remain valid, only blocks $(s{+}1){:}t$ are evaluated over the context, in parallel, rather than re-prefilling it. The other three avoid this pass and instead alter the attention pattern: \emph{masking} lets each newly activated head attend only to tokens for which it has entries, \emph{copying} substitutes the entries of an existing head, and \emph{imputing} estimates the missing entries from the token's available prefix through the head's own projections (\cref{app:kv-cache}).

A token processed at a small budget acts as a summary of itself for the larger heads that later attend to it. This can be thought of as KV-cache compression during inference: the smaller widths act as the compressor, and their entries are exactly those the full model would have shared, so nothing has to be reconciled afterward (\Cref{app:kv-cache}).

\subsection{Progressive inference}
\label{sec:cascade}
\vspace{-0.5em}
\begin{algorithm}[H]
\caption{Cascading inference for one token}
\label{alg:progressive}
\begin{algorithmic}[1]
\REQUIRE token input, KV cache, stopping rule $\textsc{Accept}$, budgets $1{:}K$
\STATE $s \gets 1$; compute block $1$ through all layers; $z_1 \gets b + A_1 h_1^L$
\WHILE{$s < K$ \AND $\neg\,\textsc{Accept}(z_s)$}
  \STATE compute block $s{+}1$ through all layers, reusing blocks $1{:}s$ and the cache \COMMENT{no recomputation}
  \STATE $z_{s+1} \gets z_s + A_{s+1} h_{s+1}^L$; \quad $s \gets s + 1$ \COMMENT{cumulative logits, \cref{eq:cumulative-logits}}
\ENDWHILE
\STATE write blocks $1{:}s$ of this token to the cache
\RETURN $\operatorname{softmax}(z_s)$
\end{algorithmic}
\end{algorithm}
\vspace{-1em}
\method{} starts each token at the smallest width and adds blocks until the
stopping policy accepts its prediction or full width is reached. Previously
computed blocks are reused (Sec.~\ref{app:cascade-cost}). In \cref{alg:progressive}, $\operatorname{Accept}(z_s)$
denotes the complete stopping policy, including the MLP and its threshold. The
policy also accesses the computed hidden features $h_{1:s}^L$ and cached
observations from earlier widths.

\paragraph{Router training.}\label{router} We train one small MLP per non-final width to estimate the difference between
its predictive distribution and that of the full model:
\[
  d_s(x)=D_{\mathrm{KL}}\!\left(
    p_K(\cdot\mid x)\,\|\,p_s(\cdot\mid x)
  \right),
  \qquad p_s=\operatorname{softmax}(z_s).
\]
The MLP uses the hidden features, confidence summaries, and prediction features
available at that width. It accepts predictions with sufficiently small
estimated discrepancy; otherwise, computation continues. Full width is always
accepted. During policy training, the language model remains frozen. A full-width pass
provides all prefix outputs and their KL targets. We fit the MLPs using mean
squared error on standardized KL or log-KL targets. A separate calibration
split selects model settings and per-width thresholds to minimize mean KL
under a chosen mean FLOP budget.

\subsection{Is \method{} expressive enough?}
\label{sec:svd}

The blocks of \cref{sec:method:blocks} zero out a large part of every hidden weight, and a natural worry is whether much is lost. Truncated SVD provides a test case: it is the prototypical progressive representation, since increasing the rank appends a component and leaves the others untouched, and it is optimal at every rank. A prefix-invariant network should therefore be able to recover it, and it does. The rank-$r$ truncation of $A^\star=U\Sigma V^\top$ is exactly a width-$r$ prefix of one factorization, a dense map $\Sigma V^\top$ into the singular basis followed by a dense readout $U$ whose columns accumulate as in \cref{eq:cumulative-logits}. This is \method{} with one hidden layer, and adding depth changes nothing:

\begin{theorem}[Depth preserves the SVD chain]
\label{thm:deep-linear-svd}
For any depth and any set of widths, there is a prefix-invariant linear network with dense endpoints whose full width computes $A^\star$ and whose width-$r$ prefix computes the optimal rank-$r$ truncation, simultaneously for every deployed $r$.
\end{theorem}
The block-triangular constraint thus incurs no cost in the linear case: the class contains the optimal predictor for all widths and depths simultaneously.
Beyond the linear case, no such simultaneous optimality can be expected, since the per-width objectives are no longer orthogonal as they are for truncated SVD. What the construction does guarantee is the structural counterpart: later blocks never alter what earlier ones compute, so training a larger submodel cannot degrade a smaller one through the forward pass, which is the interference that many-in-one models must otherwise counteract with sampling schedules and distillation~\citep{yu2019universally,devvrit2024matformer,zaccone2026flexrank}.

\section{Experiments}
\label{sec:experiments}

\textbf{Training.}
\label{sec:experiments:training}
We consider \llama{}, \nemotron{}, and \qwen{} architectures on a data mixture adapted from SmolLM2 \citep{allal2025smollm}. We fix a Token-per-Parameter (TPP) ratio of $30$, and extend it to $90$ when evaluating \llama{} on open-ended text generation (see below).
The ablation study uses a \gpt{} model, trained on FineWebEdu-$100$BT \citep{penedo2024the} (more details in \cref{sec:appendix:experiments:training}).

\textbf{Evaluation.} Our evaluation focuses on elastic and cascading inference, and compares \method{} against Slimmable networks, the closest width-based state-of-the-art method. For additional context, we extend the comparison to include CaLM and Layer-Skip as representatives of early-exit methods. 
For both, we evaluate the mean zero-shot accuracy of models across commonsense datasets commonly used in the literature (see Sec~\ref{sec:appendix:experiments:benchmarks}, using the lm-eval-harness tool \citep{eval-harness}. Cascading inference uses a per-model router trained on FineWebEdu with KL policies.
For \gpt{}, we show results on the evaluation loss on a held-out split of the pretraining dataset, as is common in the literature.

\begin{revised}
\textbf{Realistic Speedup Measurement.}
We measure the decode time of a \llama{} architecture with optimized kernels, comparing \method{} to dense and a parameter-matched sparse baselines (additional details in \cref{sec:appendix:experiments:generation_speed}).
We evaluate the quality–compute trade-off of progressive inference of generated sequences using an LLM-as-a-judge evaluation as a proxy for generation quality. We use the first-turn subset of \mbox{MT-Bench} \citep{Bai_2024} on instruction-tuned models~(\cref{sec:appendix:experiments:instruction_tuning,sec:appendix:experiments:evaluation}). 
The router used for cascading inference is calibrated on the instruction tuning dataset.
The same setting is used to investigate the impact of different KV cache approximation strategies compared to exact backfilling.
\end{revised}

\subsection{Main Results}
\vspace{-0.5em}
\begin{revised}

\paragraph{Performance on Downstream Tasks.}
\begin{figure}[t]
\centering
\definecolor{cascadeMethod}{HTML}{C44E52}
\definecolor{cascadeSlimmable}{HTML}{4C72B0}
\definecolor{cascadeDense}{HTML}{444444}
\begin{tikzpicture}
\begin{groupplot}[
  paper triple,
  scale only axis,
  width=\dimexpr(\linewidth-1.7cm)/3\relax,
  group style={
    group size=3 by 1,
    horizontal sep=0.25cm,
    y descriptions at=edge left,
  },
  ymin=38,ymax=51,
  ytick={38,40,42,44,46,48,50},
  scaled ticks=false,
]
\nextgroupplot[
  title={Llama-3.2-1B},
  xmin=0,xmax=4.7,xtick={0,1,2,3,4},
  ylabel={Mean score (\%)},
  legend to name=klbenchmarklegend,
  legend columns=-1,
  legend style={
    /tikz/every even column/.append style={column sep=0.1cm}
  },
]
\addlegendimage{empty legend}
\addlegendentry{\textbf{Cascade:}}
\addlegendimage{cascadeMethod,solid,no marks}
\addlegendentry{\method{}}
\addlegendimage{cascadeSlimmable,solid,no marks}
\addlegendentry{Slimmable}
\addlegendimage{empty legend}
\addlegendentry{\quad\textbf{Fixed widths:}}
\addlegendimage{cascadeMethod,only marks,mark=*}
\addlegendentry{\method{}}
\addlegendimage{cascadeSlimmable,only marks,mark=triangle*}
\addlegendentry{Slimmable}
\addlegendimage{cascadeDense,only marks,mark=diamond*}
\addlegendentry{Dense}

\addplot[cascadeMethod,solid,no marks] coordinates {
    (0.472413172,41.334228734) (0.570375229,42.445963109) (0.692464869,43.546529320)
    (0.824409821,44.545055930) (0.945984713,45.019253245) (1.054426213,45.359443943)
    (1.154722389,45.942565562) (1.242762518,46.061471840) (1.342010840,46.413377829)
    (1.447917494,46.601077409) (1.559801344,46.672583261) (1.648591257,46.693317065)
    (1.741389370,46.983229632) (1.803838382,47.046546661) (1.837790346,47.067608167)
};
\addplot[cascadeMethod,only marks,mark=*] coordinates {
    (0.472413172,41.334228734) (1.218504680,45.940261620) (1.837533834,47.067608167)
};
\addplot[cascadeSlimmable,solid,no marks] coordinates {
    (0.472413172,41.680166010) (0.833421335,42.424616482) (1.243429515,43.199521987)
    (1.583562393,43.772941883) (1.806970403,44.287554250) (2.011069849,44.289473503)
    (2.401517885,44.566356562) (2.822806823,44.523364652) (3.225280900,43.973618931)
    (3.578840537,44.331112064) (3.846444464,44.577939730) (4.068072875,44.443238201)
    (4.237915291,44.489078828) (4.353065970,44.533089029) (4.440712806,44.529769637)
};
\addplot[cascadeSlimmable,only marks,mark=triangle*] coordinates {
    (0.472413172,41.680166010) (1.492183016,44.227770824) (2.476116618,44.529769637)
};
\addplot[cascadeDense,only marks,mark=diamond*] coordinates {
    (2.476116618,48.367464880)
};

\nextgroupplot[
  title={Nemotron-Flash 1B},
  xmin=0,xmax=4.3,xtick={0,1,2,3,4},
]
\addplot[cascadeMethod,solid,no marks] coordinates {
    (0.229247047,38.896270508) (0.327840989,40.573046142) (0.448225188,42.437407095)
    (0.552442979,43.408300617) (0.648747520,43.939120599) (0.748591644,44.829429630)
    (0.849611252,44.981273816) (0.942911966,45.155987404) (1.041287527,45.236297327)
    (1.120314664,45.011050673) (1.203970328,45.388866054) (1.279723971,45.543369330)
    (1.341904696,45.722334571) (1.396720395,45.759966944) (1.432359197,45.810927114)
};
\addplot[cascadeMethod,only marks,mark=*] coordinates {
    (0.229247047,38.896270508) (0.543531151,42.766824956) (0.944437462,45.023475332)
    (1.431965981,45.810927114)
};
\addplot[cascadeSlimmable,solid,no marks] coordinates {
    (0.229247047,39.331454664) (0.598771522,42.261874162) (0.881119334,42.975090091)
    (1.193015220,43.947607982) (1.526822892,44.853441470) (1.795368647,45.153158554)
    (1.996242084,45.274733081) (2.292843733,45.485648110) (2.657077112,45.930970355)
    (3.028870615,46.044466007) (3.341359807,45.881426170) (3.599434406,46.196185479)
    (3.776660989,46.108485419) (3.914067947,46.162446052) (4.024840905,46.143637666)
};
\addplot[cascadeSlimmable,only marks,mark=triangle*] coordinates {
    (0.229247047,39.331454664) (0.631726223,43.014749423) (1.207449814,45.391099548)
    (1.956417821,46.143637666)
};
\addplot[cascadeDense,only marks,mark=diamond*] coordinates {
    (1.956417821,48.627238377)
};

\nextgroupplot[
  title={Qwen3.5-2B},
  xmin=0,xmax=8.2,xtick={0,2,4,6,8},
]
\addplot[cascadeMethod,solid,no marks] coordinates {
    (0.436366897,40.116434569) (0.632603713,42.500770076) (0.859793972,44.378423597)
    (1.043658398,45.312737669) (1.229963240,46.281606423) (1.407933016,46.984193347)
    (1.597503360,47.936374831) (1.785800743,48.162403409) (1.982737447,48.392705030)
    (2.165799046,48.416936905) (2.330319225,48.386328445) (2.476461014,48.920879543)
    (2.622364908,48.797365835) (2.720952321,48.764673917) (2.785187524,48.757774718)
};
\addplot[cascadeMethod,only marks,mark=*] coordinates {
    (0.436366897,40.116434569) (1.045896290,44.199406462) (1.828588179,48.205588878)
    (2.784442564,48.757774718)
};
\addplot[cascadeSlimmable,solid,no marks] coordinates {
    (0.436366897,41.270631504) (1.128702200,44.056614643) (1.723275287,45.622222877)
    (2.399683867,46.990816535) (2.980487089,47.461187832) (3.511072082,48.013730950)
    (4.031480922,48.313275556) (4.707749607,48.236336810) (5.345980337,48.808125420)
    (5.986256386,48.895619757) (6.523650647,49.121100012) (6.991380115,48.949348139)
    (7.351056033,49.109073350) (7.629849471,49.051635561) (7.826918889,49.084567610)
};
\addplot[cascadeSlimmable,only marks,mark=triangle*] coordinates {
    (0.436366897,41.270631504) (1.219058786,45.240076504) (2.348075667,48.209525913)
    (3.823417540,49.084567610)
};
\addplot[cascadeDense,only marks,mark=diamond*] coordinates {
    (3.823417540,50.342726906)
};
\end{groupplot}

\node[anchor=north,font=\PlotFont,inner sep=0pt]
  at ([yshift=-0.45cm]group c2r1.south)
  {Mean decode GFLOPs/token};

\node[anchor=north,inner sep=0pt]
  at ([yshift=-\PlotLegendGap]group c2r1.south)
  {\pgfplotslegendfromname{klbenchmarklegend}};
\end{tikzpicture}
\caption{
\textbf{\method{} enables more compute-efficient cascading inference than Slimmable.}
Across most of the shared accuracy range, \method{} matches scores with fewer decode FLOPs, while recomputation limits Slimmable’s gains. Curves denote cascades and markers denote fixed models. Scores and costs are averaged equally over six tasks (per-task results presented in \cref{fig:cascading-per-dataset}
).
}
\label{fig:cascading_benchmarks}
\end{figure}
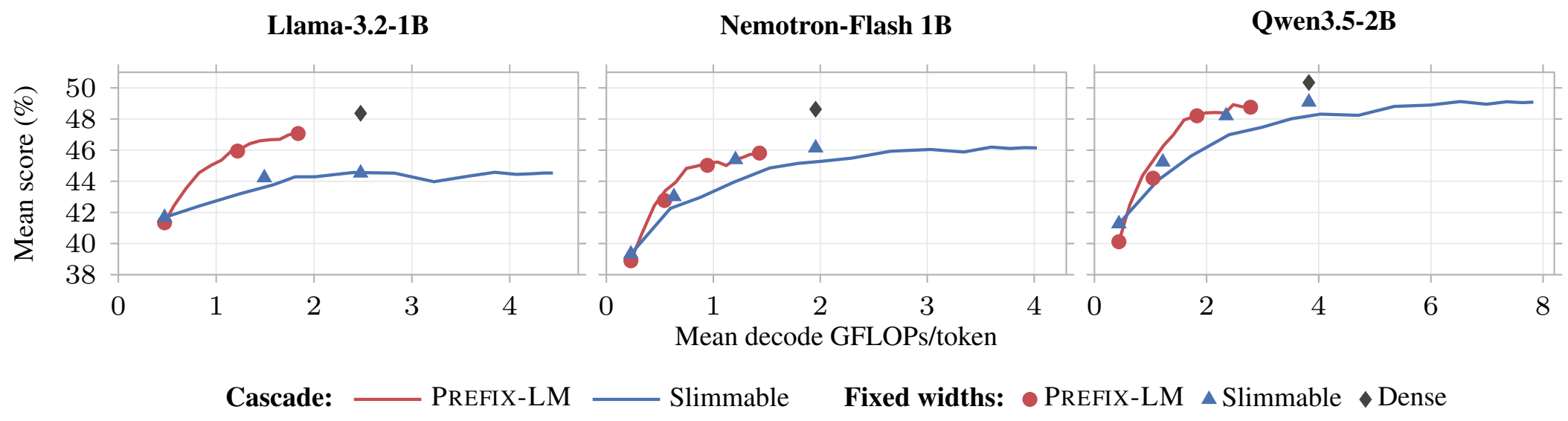
In \cref{fig:cascading_benchmarks}, we compare \method{} against Slimmable networks and evaluate the trade-off between theoretical FLOP savings and accuracy on commonsense reasoning tasks. Cascading inference with \method{} achieves higher accuracy than Slimmable baselines at the same FLOPs budgets. The standalone models achieve similar or higher accuracy to the Slimmable models on their own, and the cascade's efficiency gains are enabled by the reusability of computations across \method{} submodels. In contrast, cascading inference does not benefit Slimmable models equally, as they must recompute, sacrificing FLOP savings.

\textbf{Throughput Speedups of Cascading Inference.}
\input{figures/kv-and-speedup-plots}
\Cref{fig:mtbench_a100_speedup_matched20} shows that ideal FLOP savings translate into wall-clock speedups. We measure throughput and relate to the original dense architecture, using in both case optimized kernels, in particular for \method{} supporting cascading inference (\cref{app:throughput}). 
To confirm that these speedups are the result of cascading inference with \method{}, we provide two additional speed baselines: parameter-matched and sparsity-matched (\cref{sec:appendix:experiments:generation_speed}).
The former compares \method{} with a parameter-matched dense model, which is 6.85\% faster, while the latter compares with a randomly sparse model of the same size, which is 40\% slower than its dense counterpart. This suggests that the sparsity pattern of \method{} is well aligned with GPU hardware, contributing to its practical speedups~\citep{kwon2023efficient}.

\textbf{Tradeoffs in Historical KV Cache Approximation.}
The results in \cref{fig:mtbench_kv_performance_cost} show that cascading inference remains beneficial when generating sequences, reporting the quality and cost tradeoffs of the different KV cache approximation strategies detailed in \cref{sec:method:cache}. Tokens generated by smaller submodels do not populate the KV cache of attention heads belonging to larger submodels, and as a result, efficient solutions should approximate skipped entries as opposed to \textit{backfilling} them.
\Needspace{20\baselineskip}
\begin{wrapfigure}{R}{0.38\textwidth}
\centering
\captionsetup{font=small,skip=3pt}
\definecolor{klPrefix}{HTML}{C44E52}
\definecolor{klCalm}{HTML}{55A868}
\definecolor{klSkip}{HTML}{DD8452}
\begin{tikzpicture}
\begin{axis}[
  paper axis,width=4cm,height=\PlotHeight,
  xmin=0.3,xmax=4.2,xtick={0.5,1,1.5,2,2.5,3,3.5,4},
  ymin=35.5,ymax=48.5,ytick={36,38,40,42,44,46,48},
  xlabel={Mean ideal decode GFLOPs/token},ylabel={Mean score (\%)},
  scaled ticks=false,
  legend style={legend columns=2,at={(0.5,-0.40)},anchor=north,yshift=\PlotLegendGap-0.2cm},
]
\addlegendimage{klPrefix,solid,mark=diamond*}\addlegendentry{\method{}}
\addlegendimage{klCalm,solid,mark=*}\addlegendentry{CALM}
\addlegendimage{klSkip,solid,mark=triangle*}\addlegendentry{LayerSkip}
\addlegendimage{black!70,only marks,mark=star}\addlegendentry{Dense}
\addplot[klPrefix,solid,no marks,forget plot] coordinates {
  (0.473505,40.588065) (0.573784,41.747882) (0.697432,43.118849)
  (0.830832,43.728266) (0.953597,44.183478) (1.059226,44.531539)
  (1.157427,45.220359) (1.245883,45.593410) (1.347528,45.839170)
  (1.453366,46.034334) (1.561532,46.164087) (1.651044,45.943340)
  (1.743157,46.469275) (1.805077,46.562813) (1.840769,46.576860)
};
\addplot[klPrefix,only marks,mark=diamond*,forget plot] coordinates {
  (0.472418,40.588065) (1.218514,45.041756) (1.837546,46.576860)
};
\addplot[klCalm,solid,no marks,forget plot] coordinates {
  (1.015432,39.270171) (1.217875,40.754959) (1.522096,43.201358)
  (1.801193,44.282126) (2.034724,45.443243) (2.258129,45.640874)
  (2.504190,46.535503) (2.753644,47.250499) (2.980567,47.110636)
  (3.195649,47.092974) (3.421008,47.188055) (3.644314,47.291744)
  (3.847671,47.440502) (3.984663,47.408298) (4.059350,47.409958)
};
\addplot[klCalm,only marks,mark=*,forget plot] coordinates {
  (1.013035,39.270171) (1.500733,44.761213)
  (1.988431,46.889242) (2.476129,47.409958)
};
\addplot[klSkip,solid,no marks,forget plot] coordinates {
  (1.015432,36.091914) (1.192076,37.798885) (1.460127,39.525228)
  (1.744829,41.069436) (1.972982,41.964109) (2.192815,43.258079)
  (2.423316,44.574421) (2.683092,45.271817) (2.928950,45.384366)
  (3.150888,46.249414) (3.339103,46.314164) (3.629992,46.790043)
  (3.840649,46.721126) (3.976564,46.782537) (4.059350,46.732153)
};
\addplot[klSkip,only marks,mark=triangle*,forget plot] coordinates {
  (1.013035,36.091914) (1.500733,41.384195)
  (1.988431,45.876392) (2.476129,46.732153)
};
\addplot[black!70,only marks,mark=star,mark size=2.5pt,forget plot]
  coordinates {(2.476129,47.880243)};
\end{axis}
\end{tikzpicture}
\caption{\textbf{Width-based cascading inference reaches comparable accuracy with less computation than depth-based cascading inference.}
Under the same KL policy. CALM reaches a higher maximum score, so the advantage concerns
computation at comparable accuracy.}
\label{fig:depth-cascades}
\vspace{-4em}
\end{wrapfigure}
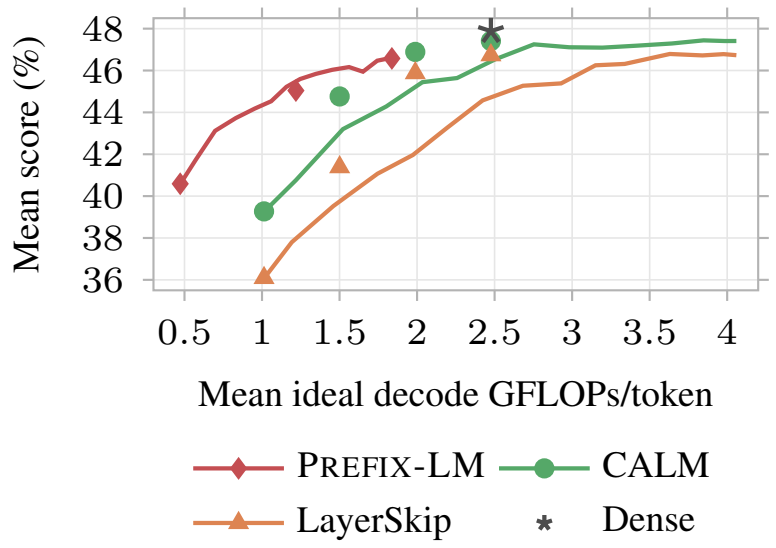
\textit{Backfilling} achieves high quality, but any efficiency gains by generating a token with a smaller submodel are outweighed by the need to recompute the KV cache of skipped entries. \textit{Masking} emerges as an approximation strategy that achieves high quality while saving FLOPs, requiring only 70.2\% of the FLOPs of the fixed-width medium model to match its quality, a $1.42\times$ FLOPs speedup.

\textbf{Comparison with Early Exiting.}
\label{app:depth-cascades}
We finally compare width-elastic \method{} with early-exit methods such as LayerSkip and CALM on downstream tasks in \cref{fig:depth-cascades}. Based on the results, \method{} achieves higher accuracy within the same FLOP budget \citep{schuster2022confident,elhoushi2024layerskip}. We note that width-wise and depth-wise adaptivity operate on different axes, so they could be combined, resulting in a form of joint adaptivity. 

\vspace{1em}
\subsection{Ablations}

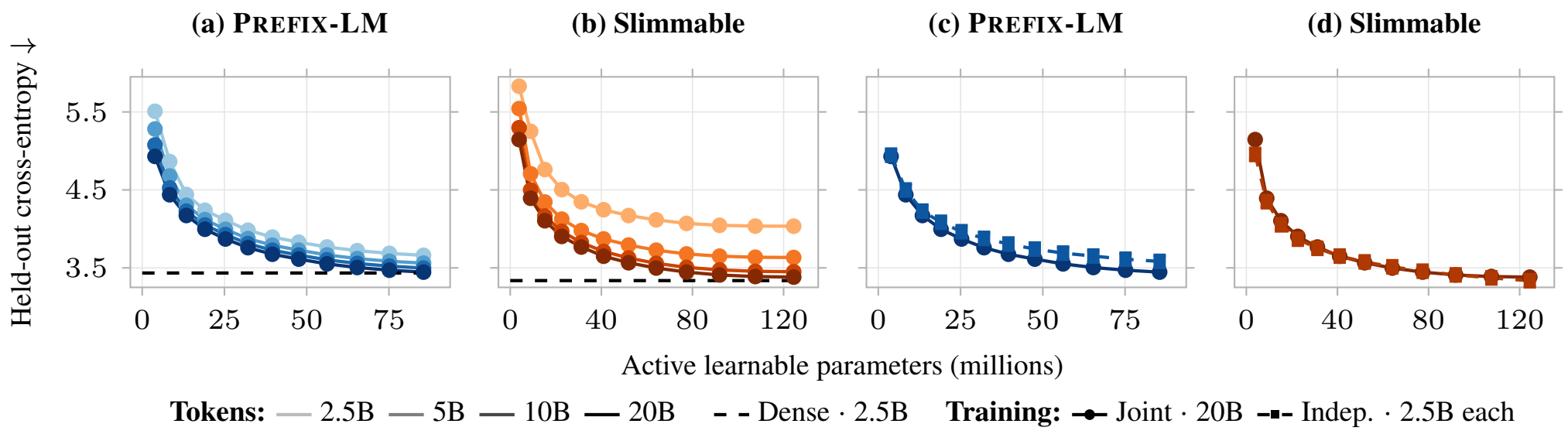
\begin{figure}[b]
\begingroup%

\definecolor{pfB25}{HTML}{9cc9e1}%
\definecolor{pfO25}{HTML}{fdad69}%
\definecolor{pfG25}{HTML}{bbbbbb}%
\definecolor{pfB5}{HTML}{519ccc}%
\definecolor{pfO5}{HTML}{f57622}%
\definecolor{pfG5}{HTML}{7f7f7f}%
\definecolor{pfB10}{HTML}{1b69af}%
\definecolor{pfO10}{HTML}{cd4401}%
\definecolor{pfG10}{HTML}{474747}%
\definecolor{pfB20}{HTML}{083573}%
\definecolor{pfO20}{HTML}{852904}%
\definecolor{pfG20}{HTML}{060606}%
\definecolor{pfB30}{HTML}{5ba3d0}%
\definecolor{pfB90}{HTML}{083c7d}%
\definecolor{pfBkey}{HTML}{0d57a1}%
\definecolor{pfO30}{HTML}{f87f2c}%
\definecolor{pfO90}{HTML}{8e2d04}%
\definecolor{pfOkey}{HTML}{b03903}%
\definecolor{pfArchitecture}{HTML}{2166AC}%
\definecolor{pfParameters}{HTML}{E69F00}%

\newcommand{\pfkey}[3]{\tikz[baseline=-.5ex]{%
  \draw[#1,#2,line width=.8pt] (0,0)--(.30,0);%
  \ifcase#3\or%
    \fill[#1] (.15,0) circle[radius=1.3pt];\or%
    \fill[#1] (.115,-.035) rectangle (.185,.035);\or%
    \draw[draw=#1,fill=white,line width=.6pt] (.115,-.035) rectangle (.185,.035);\or%
    \fill[#1] (.15,.05)--(.20,0)--(.15,-.05)--(.10,0)--cycle;%
  \fi}}%
\newcommand{\pfbox}[1]{\tikz[baseline=-.5ex]{\fill[#1,opacity=.8] (0,-.025) rectangle (.30,.075);}}%
\pgfplotsset{pfaxis/.style={paper triple,
  scaled ticks=false,/pgf/number format/1000 sep={},
  title style={align=center},enlarge x limits=.04}}
\pgfplotstableread[row sep=\\,col sep=space]{%
x j25 j5 j10 j20 ind \\%
3.88192 5.50876850598 5.28105868552 5.07725753234 4.9299475682 4.945836364 \\%
8.353664 4.86378689604 4.67813063424 4.52457990448 4.43756653867 4.49909557337 \\%
13.415232 4.44139235051 4.30250438349 4.22697304612 4.17186619226 4.22752064213 \\%
19.066624 4.23941138856 4.11904100671 4.05001817802 3.99521363945 4.08472597083 \\%
25.30784 4.10997999519 3.99445981587 3.92659895512 3.87062204941 3.9635786083 \\%
32.13888 3.98290824608 3.87553023132 3.81098175765 3.75922843865 3.878167166 \\%
39.559744 3.89490820182 3.79051584664 3.72642909547 3.67487605487 3.80596476415 \\%
47.570432 3.82920953103 3.72727599932 3.66432828994 3.61263848065 3.74136039789 \\%
56.170944 3.76826919254 3.66639280357 3.60333132906 3.5512012151 3.68940016443 \\%
65.36128 3.72100128195 3.61916666172 3.55682189438 3.50479459248 3.65123649567 \\%
75.14144 3.68588135266 3.58489528649 3.52244199083 3.47031364654 3.61102970558 \\%
85.511424 3.66109703014 3.56050390259 3.4974453451 3.44540534108 3.58300761142 \\%
}\pfTri%
\pgfplotstableread[row sep=\\,col sep=space]{%
x j25 j5 j10 j20 ind \\%
3.88192 5.82886571371 5.54287242104 5.29760545546 5.14538991837 4.95460908494 \\%
8.943488 5.25062746685 4.70692716929 4.49869442296 4.39262771713 4.34872073554 \\%
15.184704 4.76091904162 4.34464206855 4.16506979089 4.10470671303 4.05238258257 \\%
22.605568 4.50353268389 4.12498024843 3.96817960853 3.90348510101 3.8680711861 \\%
31.20608 4.34731518132 3.97799160307 3.82517376063 3.76733023441 3.75043054829 \\%
40.98624 4.24528433936 3.8711994283 3.7107013742 3.64944848146 3.65102369593 \\%
51.946048 4.1710452607 3.79292496309 3.62690299415 3.56553364556 3.57682729482 \\%
64.085504 4.11584858786 3.72681798724 3.55859877534 3.49811592114 3.51345988905 \\%
77.404608 4.07014379655 3.67896045246 3.50654671811 3.44558667175 3.45762792754 \\%
91.90336 4.04601850396 3.64999847569 3.47401475335 3.40895290202 3.40995331064 \\%
107.58176 4.03583720417 3.63629393021 3.45577742476 3.38891028888 3.37151938165 \\%
124.439808 4.03452559585 3.63310082192 3.45073447045 3.38138935076 3.33473046527 \\%
}\pfSlim%

\begin{tikzpicture}
\pgfmathsetlengthmacro{\pfScalingGap}{0.55*(\PlotGap)}
\pgfmathsetlengthmacro{\pfQuadWidth}{(\linewidth-1.8cm-3*\pfScalingGap)/4}
\begin{groupplot}[pfaxis,group style={group size=4 by 1,horizontal sep=\pfScalingGap},
  width=\pfQuadWidth,ymin=3.25,ymax=5.95,ytick={3.5,4.5,5.5}]
\nextgroupplot[title={(a) \method{}},xmin=0,xmax=90,xtick={0,25,50,75},ylabel={Held-out cross-entropy $\downarrow$}]
\addplot[color=pfB25,mark=*] table[x=x,y=j25] {\pfTri};
\addplot[color=pfB5,mark=*] table[x=x,y=j5] {\pfTri};
\addplot[color=pfB10,mark=*] table[x=x,y=j10] {\pfTri};
\addplot[color=pfB20,mark=*] table[x=x,y=j20] {\pfTri};
\addplot[black,dashed,no marks] coordinates {(0,3.43416452852) (90,3.43416452852)};
\nextgroupplot[title={(b) Slimmable},xmin=0,xmax=130,xtick={0,40,80,120},yticklabels={}]
\addplot[color=pfO25,mark=*] table[x=x,y=j25] {\pfSlim};
\addplot[color=pfO5,mark=*] table[x=x,y=j5] {\pfSlim};
\addplot[color=pfO10,mark=*] table[x=x,y=j10] {\pfSlim};
\addplot[color=pfO20,mark=*] table[x=x,y=j20] {\pfSlim};
\addplot[black,dashed,no marks] coordinates {(0,3.33682939602) (130,3.33682939602)};
\nextgroupplot[title={(c) \method{}},xmin=0,xmax=90,xtick={0,25,50,75},yticklabels={}]
\addplot[color=pfB20,mark=*] table[x=x,y=j20] {\pfTri};
\addplot[color=pfBkey,mark=square*,dashed] table[x=x,y=ind] {\pfTri};
\nextgroupplot[title={(d) Slimmable},xmin=0,xmax=130,xtick={0,40,80,120},yticklabels={}]
\addplot[color=pfO20,mark=*] table[x=x,y=j20] {\pfSlim};
\addplot[color=pfOkey,mark=square*,dashed] table[x=x,y=ind] {\pfSlim};

\end{groupplot}
\node[font=\PlotFont,anchor=north] at ($(group c1r1.south west)!.5!(group c4r1.south east)+(0,-.45cm)$) {Active learnable parameters (millions)};
\node[font=\PlotFont,anchor=north] at ($(group c1r1.south west)!.5!(group c4r1.south east)+(0,-\PlotLegendGap)$) {%
\textbf{Tokens:} \pfkey{pfG25}{solid}{0} 2.5B\hspace{.4em}\pfkey{pfG5}{solid}{0} 5B\hspace{.4em}\pfkey{pfG10}{solid}{0} 10B\hspace{.4em}\pfkey{pfG20}{solid}{0} 20B
\quad \pfkey{black}{dashed}{0} Dense $\cdot$ 2.5B
\quad \textbf{Training:} \pfkey{black}{solid}{1} Joint $\cdot$ 20B\hspace{.4em}\pfkey{black}{dashed}{2} Indep. $\cdot$ 2.5B each};
\end{tikzpicture}
\endgroup
    \caption{
    \textbf{\method{} scales better than Slimmable across training token budget:} performance scaling across submodels for \method{} (a) and Slimmable (b). Comparing to non-elastic training, \method{} surpasses independently trained models with less overall training budget (c).
    }
    \label{fig:abl:scaling}
\end{figure}
\textbf{Scaling with training tokens.}
\Cref{fig:abl:scaling} compares \method{} and Slimmable, jointly training 12 nested submodels at evenly-spaced width increments. We analyze the scaling at $2.5$B, $5$B, $10$B, and $20$B tokens.
Plots (a--b) show improvements across widths as training increases, with the largest submodels approaching dense references matched to their parameter counts and trained for $2.5$B tokens.
Plots (c--d) compare joint and independent training, similarly as in~\cite[Fig.~5]{zaccone2026flexrank}. In model elasticity, the ability to instantiate smaller models without retraining from scratch~\citep{yu2019slimmable} motivates additional training, provided that the shared effort remains competitive with training those models separately~\citep{cai2020once}.
For \method{}, all the submodels trained jointly for $20$B tokens outperform their counterparts trained independently for $2.5$B tokens each, indicating successful elastic training and higher cost-effectiveness optimization than Slimmable.

\textbf{Architectural and parameter-count costs.}
\input{figures/abl_triangular}
\Cref{fig:abl:triag_analysis}(a) shows the loss gap decomposition between \method{} and Slimmable at the same hidden width, averaged over the four training budgets
(more details on the procedure in \cref{abb:decomp_arch}).
The ratio is roughly $60\%$ architectural and $40\%$ parameter-count-related, and remains stable across widths and budgets (see \cref{fig:app:triag_analysis}).
\Cref{fig:abl:triag_analysis}(b) quantifies the quality cost of choosing triangular connectivity over an unconstrained dense architecture at the same parameter and training budgets. This cost results in a $\approx 5\%$ higher held-out cross-entropy, indicating that the structure for prefix invariance comes with a modest regret.

\textbf{Sample efficiency of joint training.}
\Cref{fig:abl:triag_analysis}(c) compares jointly and independently trained submodels within each architecture at equal token budgets, showing a smaller joint-training penalty for \method{} than for Slimmable. This is motivated by prefix invariance joint submodel optimization in a single shared forward pass, making it cost-effective to optimize every submodel on every batch.

\end{revised}

\section{Related Work}
\label{sec:related}
\vspace{-1em}
\textbf{Elastic models and nested representations.}
Elastic models make one set of weights usable at several compute budgets. Slimmable networks do so by varying channel width \citep{yu2019slimmable,yu2019universally}, and once-for-all networks allow broader architecture choices \citep{cai2020once}. For language models, MatFormer nests feed-forward blocks \citep{devvrit2024matformer}, while Flextron, LlamaFlex, Star Elastic, and FlexRank offer other forms of shared, variable-capacity models \citep{cai2024flextron,cai2025llamaflex,taghibakhshi2026starelastic,zaccone2026flexrank}. These approaches reduce the cost of maintaining multiple model sizes. However, sharing weights does not imply representation reuse. Increasing the budget can change activations that the smaller model has already produced. \method{} addresses this distinction by preserving those activations at every layer. This also differs from Matryoshka representations and nested dropout, which make truncated representations useful without requiring separately executed widths to agree throughout the network \citep{kusupati2022matryoshka,rippel2014nested}.

\textbf{Adaptive and progressive inference.}
Once several budgets are available, inference can choose where to spend them. Routing selects a model for a request or token \citep{ong2025routellm,zheng2024citer}; cascading inference tries a cheaper model first and escalates when needed \citep{narasimhan2025faster}. Speculative decoding uses a stronger model to verify cheap drafts \citep{leviathan2023faster}. The benefit of ``escalating'' depends on how much of the first model's work survives the switch.
Progressive inference seeks precisely this reuse as a prediction is refined \citep{laskaridis2020spinn,laskaridis2020hapi,zhang2018progressiveneuralnetworksimage}. In language generation, CALM and LayerSkip pursue it along depth, while managing the KV states of tokens that exited early \citep{schuster2022confident,elhoushi2024layerskip,bae2023fast}. \method{} instead adds width while retaining full depth. 
With a shared token history and attention mask, \method{} preserves existing KV entries exactly across widths; only missing blocks need to be computed for full-context attention.

\textbf{Triangular connections.}
Preserving earlier work as width grows requires controlling which channels can influence each other. Doubly Nested Networks use channel-causal convolutions to support width and depth slicing \citep{kim2018doubly}, and Any-Width Networks use lower-triangular weights in adjustable-width CNNs \citep{vu2020anywidth}. Unlike prior work, focusing on small CNNs, we formulate the underlying requirement as prefix invariance and enforce it across Transformer operations, including normalization, attention, and residual paths. Our work is the first to extend logits and KV caches incrementally during autoregressive generation, giving true progressive inference for language models.

\vspace{-1em}
\section{Conclusion}
\vspace{-1em}
We introduced \method{}, a Transformer whose submodels are prefixes of one another at every layer. This prefix invariance lets additional width refine rather than replace earlier computation, so predictions, hidden states, and KV caches are reused exactly across budgets. The class recovers truncated SVD at any depth in the linear case, and in practice, a learned router cascades billion-parameter models to near full-model agreement at a fraction of the cost, with the savings carrying through to decoding throughput. Investigating better cascading policies and combining width-wise refinement with depth-wise early exit are natural next steps.

\subsection*{AI use statement}

We used generative AI tools to assist with implementing and debugging
\method{}'s cascading policies and evaluation pipelines, and with
optimizing inference code for decoding throughput. These tools also
supported analysis of agreement--compute tradeoffs, discussion of
theoretical arguments, figure preparation, and manuscript drafting
and editing. Implementation checks included automated tests and
comparisons with reference outputs. The authors take responsibility
for the final methodology, claims, experimental results, and artifacts.

\section*{Funding}
Riccardo Zaccone declares that financial support was received for the research, authorship, and/or publication of this article. This study was carried out within the project FAIR - Future Artificial Intelligence Research - and received funding from the European Union Next-GenerationEU [PIANO NAZIONALE DI RIPRESA E RESILIENZA (PNRR) – MISSIONE 4 COMPONENTE 2, INVESTIMENTO 1.3 – D.D. 1555 11/10/2022, PE00000013 - CUP: E13C22001800001]. This manuscript reflects only the authors’ views and opinions, neither the European Union nor the European Commission can be considered responsible for them. 
A part of the computational resources for this work was provided by hpc@polito, which is a Project of Academic Computing within the Department of Control and Computer Engineering at the Politecnico di Torino (\href{http://www.hpc.polito.it}{\texttt{http://www.hpc.polito.it}}). We acknowledge the CINECA award under the ISCRA initiative for the availability of high-performance computing resources.
We are grateful for the support of the National Science Foundation (NSF) (CCF2106707), the Defense Advanced Research Projects Agency (DARPA Young Faculty Award), and the Wisconsin Alumni Research Foundation (WARF).

\bibliography{iclr2027_conference}

\begin{thebibliography}{37}
\providecommand{\natexlab}[1]{#1}
\providecommand{\url}[1]{\texttt{#1}}
\expandafter\ifx\csname urlstyle\endcsname\relax
  \providecommand{\doi}[1]{doi: #1}\else
  \providecommand{\doi}{doi: \begingroup \urlstyle{rm}\Url}\fi

\bibitem[Ainslie et~al.(2023)Ainslie, Lee-Thorp, de~Jong, Zemlyanskiy, Lebron,
  and Sanghai]{ainslie2023gqa}
Joshua Ainslie, James Lee-Thorp, Michiel de~Jong, Yury Zemlyanskiy, Federico
  Lebron, and Sumit Sanghai.
\newblock {GQA}: Training generalized multi-query transformer models from
  multi-head checkpoints.
\newblock In \emph{The 2023 Conference on Empirical Methods in Natural Language
  Processing}, 2023.
\newblock URL \url{https://openreview.net/forum?id=hmOwOZWzYE}.

\bibitem[allal et~al.(2025)allal, Lozhkov, Bakouch, Blazquez, Penedo, Tunstall,
  Marafioti, Lajar{\'\i}n, Kydl{\'\i}{\v{c}}ek, Srivastav, Lochner, Fahlgren,
  NGUYEN, Burtenshaw, Fourrier, Zhao, Larcher, Morlon, Zakka, Raffel, Werra,
  and Wolf]{allal2025smollm}
Loubna~Ben allal, Anton Lozhkov, Elie Bakouch, Gabriel~Martin Blazquez,
  Guilherme Penedo, Lewis Tunstall, Andr{\'e}s Marafioti, Agust{\'\i}n~Piqueres
  Lajar{\'\i}n, Hynek Kydl{\'\i}{\v{c}}ek, Vaibhav Srivastav, Joshua Lochner,
  Caleb Fahlgren, Xuan~Son NGUYEN, Ben Burtenshaw, Cl{\'e}mentine Fourrier,
  Haojun Zhao, Hugo Larcher, Mathieu Morlon, Cyril Zakka, Colin Raffel,
  Leandro~Von Werra, and Thomas Wolf.
\newblock Smol{LM}2: When smol goes big {\textemdash} data-centric training of
  a fully open small language model.
\newblock In \emph{Second Conference on Language Modeling}, 2025.
\newblock URL \url{https://openreview.net/forum?id=3JiCl2A14H}.

\bibitem[Bae et~al.(2023)Bae, Ko, Song, and Yun]{bae2023fast}
Sangmin Bae, Jongwoo Ko, Hwanjun Song, and Se-Young Yun.
\newblock Fast and robust early-exiting framework for autoregressive language
  models with synchronized parallel decoding.
\newblock In Houda Bouamor, Juan Pino, and Kalika Bali (eds.),
  \emph{Proceedings of the 2023 Conference on Empirical Methods in Natural
  Language Processing}, pp.\  5910--5924, Singapore, December 2023. Association
  for Computational Linguistics.
\newblock \doi{10.18653/v1/2023.emnlp-main.362}.
\newblock URL \url{https://aclanthology.org/2023.emnlp-main.362/}.

\bibitem[Bai et~al.(2024)Bai, Liu, Bu, He, Liu, Zhou, Lin, Su, Ge, Zheng, and
  Ouyang]{Bai_2024}
Ge~Bai, Jie Liu, Xingyuan Bu, Yancheng He, Jiaheng Liu, Zhanhui Zhou, Zhuoran
  Lin, Wenbo Su, Tiezheng Ge, Bo~Zheng, and Wanli Ouyang.
\newblock Mt-bench-101: A fine-grained benchmark for evaluating large language
  models in multi-turn dialogues.
\newblock In \emph{Proceedings of the 62nd Annual Meeting of the Association
  for Computational Linguistics (Volume 1: Long Papers)}, pp.\  7421–7454.
  Association for Computational Linguistics, 2024.
\newblock \doi{10.18653/v1/2024.acl-long.401}.
\newblock URL \url{http://dx.doi.org/10.18653/v1/2024.acl-long.401}.

\bibitem[Cai et~al.(2020)Cai, Gan, Wang, Zhang, and Han]{cai2020once}
Han Cai, Chuang Gan, Tianzhe Wang, Zhekai Zhang, and Song Han.
\newblock Once-for-all: Train one network and specialize it for efficient
  deployment.
\newblock In \emph{International Conference on Learning Representations}, 2020.

\bibitem[Cai et~al.(2024)Cai, Muralidharan, Heinrich, Yin, Wang, Kautz, and
  Molchanov]{cai2024flextron}
Ruisi Cai, Saurav Muralidharan, Greg Heinrich, Hongxu Yin, Zhangyang Wang, Jan
  Kautz, and Pavlo Molchanov.
\newblock Flextron: Many-in-one flexible large language model.
\newblock In Ruslan Salakhutdinov, Zico Kolter, Katherine Heller, Adrian
  Weller, Nuria Oliver, Jonathan Scarlett, and Felix Berkenkamp (eds.),
  \emph{Proceedings of the 41st International Conference on Machine Learning},
  volume 235 of \emph{Proceedings of Machine Learning Research}, pp.\
  5298--5311. PMLR, 21--27 Jul 2024.
\newblock URL \url{https://proceedings.mlr.press/v235/cai24e.html}.

\bibitem[Cai et~al.(2025)Cai, Muralidharan, Yin, Wang, Kautz, and
  Molchanov]{cai2025llamaflex}
Ruisi Cai, Saurav Muralidharan, Hongxu Yin, Zhangyang Wang, Jan Kautz, and
  Pavlo Molchanov.
\newblock Llamaflex: Many-in-one llms via generalized pruning and weight
  sharing.
\newblock In \emph{International Conference on Learning Representations},
  volume 2025, pp.\  42250--42263, 2025.

\bibitem[Devvrit et~al.(2024)Devvrit, Kudugunta, Kusupati, Dettmers, Chen,
  Dhillon, Tsvetkov, Hajishirzi, Kakade, Farhadi, and
  Jain]{devvrit2024matformer}
Devvrit, Sneha Kudugunta, Aditya Kusupati, Tim Dettmers, Kaifeng Chen, Inderjit
  Dhillon, Yulia Tsvetkov, Hannaneh Hajishirzi, Sham Kakade, Ali Farhadi, and
  Prateek Jain.
\newblock {MatFormer}: Nested transformer for elastic inference.
\newblock In \emph{Advances in Neural Information Processing Systems}, 2024.

\bibitem[Elhoushi et~al.(2024)Elhoushi, Shrivastava, Liskovich, Hosmer, Wasti,
  Lai, Mahmoud, Acun, Agarwal, Roman, et~al.]{elhoushi2024layerskip}
Mostafa Elhoushi, Akshat Shrivastava, Diana Liskovich, Basil Hosmer, Bram
  Wasti, Liangzhen Lai, Anas Mahmoud, Bilge Acun, Saurabh Agarwal, Ahmed Roman,
  et~al.
\newblock Layerskip: Enabling early exit inference and self-speculative
  decoding.
\newblock In \emph{Proceedings of the 62nd Annual Meeting of the Association
  for Computational Linguistics (Volume 1: Long Papers)}, pp.\  12622--12642,
  2024.

\bibitem[Gao et~al.(2024)Gao, Tow, Abbasi, Biderman, Black, DiPofi, Foster,
  Golding, Hsu, Le~Noac'h, Li, McDonell, Muennighoff, Ociepa, Phang, Reynolds,
  Schoelkopf, Skowron, Sutawika, Tang, Thite, Wang, Wang, and
  Zou]{eval-harness}
Leo Gao, Jonathan Tow, Baber Abbasi, Stella Biderman, Sid Black, Anthony
  DiPofi, Charles Foster, Laurence Golding, Jeffrey Hsu, Alain Le~Noac'h,
  Haonan Li, Kyle McDonell, Niklas Muennighoff, Chris Ociepa, Jason Phang,
  Laria Reynolds, Hailey Schoelkopf, Aviya Skowron, Lintang Sutawika, Eric
  Tang, Anish Thite, Ben Wang, Kevin Wang, and Andy Zou.
\newblock The language model evaluation harness, 07 2024.
\newblock URL \url{https://zenodo.org/records/12608602}.

\bibitem[Hinton et~al.(2015)Hinton, Vinyals, and Dean]{hinton2015distilling}
Geoffrey Hinton, Oriol Vinyals, and Jeff Dean.
\newblock Distilling the knowledge in a neural network.
\newblock In \emph{NeurIPS Deep Learning and Representation Learning Workshop},
  2015.

\bibitem[Jordan et~al.(2024)Jordan, Jin, Boza, You, Cesista, Newhouse, and
  Bernstein]{jordan2024muon}
Keller Jordan, Yuchen Jin, Vlado Boza, J.~You, F.~Cesista, L.~Newhouse, and
  J.~Bernstein.
\newblock Muon: An optimizer for hidden layers in neural networks, 2024.

\bibitem[Kim et~al.(2018)Kim, Hong, Choi, and Kim]{kim2018doubly}
Jaehong Kim, Sungeun Hong, Yongseok Choi, and Jiwon Kim.
\newblock Doubly nested network for resource-efficient inference, 2018.
\newblock URL \url{https://arxiv.org/abs/1806.07568}.

\bibitem[Kusupati et~al.(2022)Kusupati, Bhatt, Rege, Wallingford, Sinha,
  Ramanujan, Howard-Snyder, Chen, Kakade, Jain, and
  Farhadi]{kusupati2022matryoshka}
Aditya Kusupati, Gantavya Bhatt, Aniket Rege, Matthew Wallingford, Aditya
  Sinha, Vivek Ramanujan, William Howard-Snyder, Kaifeng Chen, Sham Kakade,
  Prateek Jain, and Ali Farhadi.
\newblock Matryoshka representation learning.
\newblock In \emph{Proceedings of the 36th International Conference on Neural
  Information Processing Systems}, NIPS '22, Red Hook, NY, USA, 2022. Curran
  Associates Inc.
\newblock ISBN 9781713871088.

\bibitem[Kwon et~al.(2023)Kwon, Li, Zhuang, Sheng, Zheng, Yu, Gonzalez, Zhang,
  and Stoica]{kwon2023efficient}
Woosuk Kwon, Zhuohan Li, Siyuan Zhuang, Ying Sheng, Lianmin Zheng, Cody~Hao Yu,
  Joseph~E. Gonzalez, Hao Zhang, and Ion Stoica.
\newblock Efficient memory management for large language model serving with
  pagedattention.
\newblock In \emph{Proceedings of the ACM SIGOPS 29th Symposium on Operating
  Systems Principles}, 2023.

\bibitem[Laskaridis et~al.(2020{\natexlab{a}})Laskaridis, Venieris, Almeida,
  Leontiadis, and Lane]{laskaridis2020spinn}
Stefanos Laskaridis, Stylianos~I. Venieris, Mario Almeida, Ilias Leontiadis,
  and Nicholas~D. Lane.
\newblock Spinn: synergistic progressive inference of neural networks over
  device and cloud.
\newblock In \emph{Proceedings of the 26th Annual International Conference on
  Mobile Computing and Networking}, MobiCom '20, New York, NY, USA,
  2020{\natexlab{a}}. Association for Computing Machinery.
\newblock ISBN 9781450370851.
\newblock \doi{10.1145/3372224.3419194}.
\newblock URL \url{https://doi.org/10.1145/3372224.3419194}.

\bibitem[Laskaridis et~al.(2020{\natexlab{b}})Laskaridis, Venieris, Kim, and
  Lane]{laskaridis2020hapi}
Stefanos Laskaridis, Stylianos~I. Venieris, Hyeji Kim, and Nicholas~D. Lane.
\newblock Hapi: hardware-aware progressive inference.
\newblock In \emph{Proceedings of the 39th International Conference on
  Computer-Aided Design}, ICCAD '20, New York, NY, USA, 2020{\natexlab{b}}.
  Association for Computing Machinery.
\newblock ISBN 9781450380263.
\newblock \doi{10.1145/3400302.3415698}.
\newblock URL \url{https://doi.org/10.1145/3400302.3415698}.

\bibitem[Leviathan et~al.(2023)Leviathan, Kalman, and
  Matias]{leviathan2023faster}
Yaniv Leviathan, Matan Kalman, and Yossi Matias.
\newblock Fast inference from transformers via speculative decoding.
\newblock In Andreas Krause, Emma Brunskill, Kyunghyun Cho, Barbara Engelhardt,
  Sivan Sabato, and Jonathan Scarlett (eds.), \emph{Proceedings of the 40th
  International Conference on Machine Learning}, volume 202 of
  \emph{Proceedings of Machine Learning Research}, pp.\  19274--19286. PMLR,
  23--29 Jul 2023.
\newblock URL \url{https://proceedings.mlr.press/v202/leviathan23a.html}.

\bibitem[Liu et~al.(2026)Liu, Huang, Yao, Feng, Gu, Du, Li, Cheng, Jiang, Lu,
  Musuvathi, and Choukse]{liu2026droidspeak}
Yuhan Liu, Yuyang Huang, Jiayi Yao, Shaoting Feng, Zhuohan Gu, Kuntai Du,
  Hanchen Li, Yihua Cheng, Junchen Jiang, Shan Lu, Madan Musuvathi, and Esha
  Choukse.
\newblock \emph{DroidSpeak: KV cache sharing across fine-tuned model variants}.
\newblock USENIX Association, USA, 2026.
\newblock ISBN 978-1-939133-54-0.

\bibitem[Loshchilov \& Hutter(2019)Loshchilov and
  Hutter]{loshchilov2019decoupled}
Ilya Loshchilov and Frank Hutter.
\newblock Decoupled weight decay regularization.
\newblock In \emph{International Conference on Learning Representations
  (ICLR)}, 2019.
\newblock URL \url{https://openreview.net/forum?id=Bkg6RiCqY7}.

\bibitem[Mirzadeh et~al.(2019)Mirzadeh, Farajtabar, Li, Levine, Matsukawa, and
  Ghasemzadeh]{mirzadeh2019improved}
Seyed-Iman Mirzadeh, Mehrdad Farajtabar, Ang Li, Nir Levine, Akihiro Matsukawa,
  and Hassan Ghasemzadeh.
\newblock Improved knowledge distillation via teacher assistant, 2019.
\newblock URL \url{https://arxiv.org/abs/1902.03393}.

\bibitem[Narasimhan et~al.(2025)Narasimhan, Jitkrittum, Rawat, Kim, Gupta,
  Menon, and Kumar]{narasimhan2025faster}
Harikrishna Narasimhan, Wittawat Jitkrittum, Ankit~Singh Rawat, Seungyeon Kim,
  Neha Gupta, Aditya~Krishna Menon, and Sanjiv Kumar.
\newblock Faster cascades via speculative decoding.
\newblock In \emph{The Thirteenth International Conference on Learning
  Representations}, 2025.
\newblock URL \url{https://openreview.net/forum?id=vo9t20wsmd}.

\bibitem[Ong et~al.(2025)Ong, Almahairi, Wu, Chiang, Wu, Gonzalez, Kadous, and
  Stoica]{ong2025routellm}
Isaac Ong, Amjad Almahairi, Vincent Wu, Wei-Lin Chiang, Tianhao Wu, Joseph~E.
  Gonzalez, M~Waleed Kadous, and Ion Stoica.
\newblock Route{LLM}: Learning to route {LLM}s from preference data.
\newblock In \emph{The Thirteenth International Conference on Learning
  Representations}, 2025.
\newblock URL \url{https://openreview.net/forum?id=8sSqNntaMr}.

\bibitem[Penedo et~al.(2024)Penedo, Kydl{\'\i}{\v{c}}ek, allal, Lozhkov,
  Mitchell, Raffel, Werra, and Wolf]{penedo2024the}
Guilherme Penedo, Hynek Kydl{\'\i}{\v{c}}ek, Loubna~Ben allal, Anton Lozhkov,
  Margaret Mitchell, Colin Raffel, Leandro~Von Werra, and Thomas Wolf.
\newblock The fineweb datasets: Decanting the web for the finest text data at
  scale.
\newblock In \emph{The Thirty-eight Conference on Neural Information Processing
  Systems Datasets and Benchmarks Track}, 2024.
\newblock URL \url{https://openreview.net/forum?id=n6SCkn2QaG}.

\bibitem[Rippel et~al.(2014)Rippel, Gelbart, and Adams]{rippel2014nested}
Oren Rippel, Michael~A. Gelbart, and Ryan~P. Adams.
\newblock Learning ordered representations with nested dropout.
\newblock \emph{arXiv preprint arXiv:1402.0915}, 2014.

\bibitem[Safaryan et~al.(2023)Safaryan, Peste, and
  Alistarh]{safaryan2023knowledge}
Mher Safaryan, Alexandra Peste, and Dan Alistarh.
\newblock Knowledge distillation performs partial variance reduction.
\newblock In \emph{Thirty-seventh Conference on Neural Information Processing
  Systems}, 2023.
\newblock URL \url{https://openreview.net/forum?id=tzxP9Rx0LV}.

\bibitem[Schuster et~al.(2022)Schuster, Fisch, Gupta, Dehghani, Bahri, Tran,
  Tay, and Metzler]{schuster2022confident}
Tal Schuster, Adam Fisch, Jai Gupta, Mostafa Dehghani, Dara Bahri, Vinh Tran,
  Yi~Tay, and Donald Metzler.
\newblock Confident adaptive language modeling.
\newblock \emph{Advances in Neural Information Processing Systems},
  35:\penalty0 17456--17472, 2022.

\bibitem[Shazeer(2019)]{shazeer2019fast}
Noam Shazeer.
\newblock Fast transformer decoding: One write-head is all you need, 2019.
\newblock URL \url{https://arxiv.org/abs/1911.02150}.

\bibitem[Taghibakhshi et~al.(2026)Taghibakhshi, Cai, Muralidharan, Sreenivas,
  Vavre, Mahabaleshwarkar, Kartal, Liang, Chochowski, Chen,
  et~al.]{taghibakhshi2026starelastic}
Ali Taghibakhshi, Ruisi Cai, Saurav Muralidharan, Sharath~Turuvekere Sreenivas,
  Aditya Vavre, Ameya~Sunil Mahabaleshwarkar, Bilal Kartal, Sheldon Liang,
  Marcin Chochowski, Zijia Chen, et~al.
\newblock Star elastic: Many-in-one reasoning {LLM}s with efficient budget
  control.
\newblock \emph{arXiv preprint arXiv:2605.07182}, 2026.

\bibitem[Taori et~al.(2023)Taori, Gulrajani, Zhang, Dubois, Li, Guestrin,
  Liang, and Hashimoto]{alpaca}
Rohan Taori, Ishaan Gulrajani, Tianyi Zhang, Yann Dubois, Xuechen Li, Carlos
  Guestrin, Percy Liang, and Tatsunori~B. Hashimoto.
\newblock Stanford alpaca: An instruction-following llama model.
\newblock \url{https://github.com/tatsu-lab/stanford_alpaca}, 2023.

\bibitem[Tay et~al.(2022)Tay, Dehghani, Rao, Fedus, Abnar, Chung, Narang,
  Yogatama, Vaswani, and Metzler]{tay2022scale}
Yi~Tay, Mostafa Dehghani, Jinfeng Rao, William Fedus, Samira Abnar, Hyung~Won
  Chung, Sharan Narang, Dani Yogatama, Ashish Vaswani, and Donald Metzler.
\newblock Scale efficiently: Insights from pretraining and finetuning
  transformers.
\newblock In \emph{International Conference on Learning Representations}, 2022.
\newblock URL \url{https://openreview.net/forum?id=f2OYVDyfIB}.

\bibitem[Vu et~al.(2020)Vu, Eder, Price, and Frahm]{vu2020anywidth}
Thanh Vu, Marc Eder, True Price, and Jan-Michael Frahm.
\newblock Any-width networks.
\newblock In \emph{2020 IEEE/CVF Conference on Computer Vision and Pattern
  Recognition Workshops (CVPRW)}, pp.\  3018--3026, 2020.
\newblock \doi{10.1109/CVPRW50498.2020.00360}.

\bibitem[Yu \& Huang(2019)Yu and Huang]{yu2019universally}
Jiahui Yu and Thomas~S. Huang.
\newblock Universally slimmable networks and improved training techniques.
\newblock In \emph{IEEE/CVF International Conference on Computer Vision}, pp.\
  1803--1811, 2019.

\bibitem[Yu et~al.(2019)Yu, Yang, Xu, Yang, and Huang]{yu2019slimmable}
Jiahui Yu, Linjie Yang, Ning Xu, Jianchao Yang, and Thomas Huang.
\newblock Slimmable neural networks.
\newblock In \emph{International Conference on Learning Representations}, 2019.
\newblock URL \url{https://openreview.net/forum?id=H1gMCsAqY7}.

\bibitem[Zaccone et~al.(2026)Zaccone, Laskaridis, Ciccone, and
  Horv{\'a}th]{zaccone2026flexrank}
Riccardo Zaccone, Stefanos Laskaridis, Marco Ciccone, and Samuel Horv{\'a}th.
\newblock {FlexRank}: Nested low-rank knowledge decomposition for adaptive
  model deployment.
\newblock In \emph{International Conference on Machine Learning}, 2026.
\newblock URL \url{https://openreview.net/forum?id=DK0kvnNelx}.
\newblock Spotlight.

\bibitem[Zhang et~al.(2018)Zhang, Ning, Cen, Li, Zhao, Sun, and
  He]{zhang2018progressiveneuralnetworksimage}
Zhi Zhang, Guanghan Ning, Yigang Cen, Yang Li, Zhiqun Zhao, Hao Sun, and Zhihai
  He.
\newblock Progressive neural networks for image classification, 2018.
\newblock URL \url{https://arxiv.org/abs/1804.09803}.

\bibitem[Zheng et~al.(2024)Zheng, Chen, Zhang, Kundu, Li, Liu, Xing, Wang, and
  Yao]{zheng2024citer}
Wenhao Zheng, Yixiao Chen, Weitong Zhang, Souvik Kundu, Yun Li, Zhengzhong Liu,
  Eric~P. Xing, Hongyi Wang, and Huaxiu Yao.
\newblock {CITER}: Collaborative inference for efficient large language model
  decoding with token-level routing.
\newblock In \emph{Adaptive Foundation Models: Evolving AI for Personalized and
  Efficient Learning}, 2024.
\newblock URL \url{https://openreview.net/forum?id=0tSUpTUgfP}.

\end{thebibliography}
\bibliographystyle{iclr2027_conference}

\clearpage
\appendix
\section{Additional Discussion}
\label{sec:appendix:discussion}

\subsection{Efficient Execution of Distillation Loss}
\label{app:telescoping}

We exploit the fact that the output head of \method{} is cumulative because it is shared between the submodels,
\[
z_i=\sum_{j\leq i}\Delta z_j,
\qquad
p_i=\operatorname{softmax}(z_i).
\]

For equally weighted adjacent-prefix distillation with detached teachers and a final hard-label loss, the gradients reaching classifier block $i$ telescope:
\[
\nabla_{\Delta z_i}\mathcal{L}
=
\frac{1}{K}
\left[
    \sum_{j=i}^{K-1}(p_j-p_{j+1})
    +p_K-y
\right]
=
\frac{p_i-y}{K},
\]
where $y$ is the one-hot target.

For hard-label HSD, the compiled CUDA path also simplifies the backward without changing the objective. If $z_i$ is the cumulative logits at prefix $i$ and $p_i=\operatorname{softmax}(z_i)$, detached-teacher HSD produces successive logit gradients $p_i-p_{i+1}$, followed by $p_K-y$. A disjoint projection block appears in every cumulative prefix from its own prefix onward, so those gradients telescope to $p_i-y$ for that block. The implementation therefore returns the true HSD scalar while attaching the equivalent gradient from one hard-label cross-entropy per block, with preceding cumulative logits detached. This preserves the scalar and first-order parameter gradients used by the optimizer.

\subsection{KV Cache Policies During Decoding}
\label{app:kv-cache}

Let $s_i$ be the budget used to process token $i$, and let KV head $h$ belong to block $r$. Suppressing layer indices, write $\bar h_{i,1:s_i}$ for the available normalized features and $R_i$ for rotary encoding at the original position. Prompt tokens are processed at full width. Each policy specifies stored KV entries and the positions $\mathcal I_{h,t}$ over which attention is normalized.

\paragraph{Backfilling (exact reconstruction).}
Reconstruct missing historical feature blocks through all layers before evaluating a larger budget. Since blocks $1{:}s$ of those tokens are already available and remain valid, this requires evaluating only blocks $s{+}1{:}t$ over the context, rather than re-prefilling it. This policy ensures no approximation is made during ifnerence, conditioned on the sequence.

\paragraph{Masking (sparse history).}
We perform regular attention over heads whose previous entries are all filled, and sparse attention for heads that are missing some entries, depending on which tokens in the sequence cascaded to which budget:
\[
\mathcal I_{h,t}=\{i\le t:s_i\ge r\},\qquad (\tilde k_{h,i},\tilde v_{h,i})=(k_{h,i},v_{h,i}).
\]
Missing positions are excluded from the softmax.

\paragraph{Copying (using smaller KV cache).}
Let $H(s_i)$ be the number of computed KV heads, indexed from one. For each missing head, cyclically select an available head at the same layer and position:
  \[
  \phi_i(h)=1+((h-1)\bmod H(s_i)),\qquad
  (\tilde k_{h,i},\tilde v_{h,i})
  =(k_{\phi_i(h),i},v_{\phi_i(h),i}).
  \]
Computed entries remain unchanged; the heads of larger models read from the KV cache of smaller ones. Copied keys already include rotary encodings and do not need to be applied again.

\paragraph{Imputing (partial projection).}
 For each missing head in block $r>s_i$, use its own projections restricted to available features:
  \[
  \tilde k_{h,i}
  =R_iW^K_{h,1:s_i}\bar h_{i,1:s_i},\qquad
  \tilde v_{h,i}
  =W^V_{h,1:s_i}\bar h_{i,1:s_i}.
  \]
This zero-extends the already-normalized features, and computed entries remain unchanged, similarly to copying.

\subsection{Idealized Cost of Progressive Inference}
\label{app:cascade-cost}

Progressive inference computes each reached block once; restarting at every width repeats block $r$ a total of $s-r+1$ times for exit $s$. Let
\[
F_s=\sum_{r\le s}(C_r+E_r)
\]
denote the current token's non-attention backbone and incremental shared-readout FLOPs.

For $N$ layers, let $d_r,v_r$ be the residual/query and KV widths added by block $r$, and let $b$ be the bytes per cached scalar. With $t$ including the current token and $s=s_t$, define
\[
D_s=\sum_{r\le s} d_r,
\qquad
U_s=\sum_{r\le s} v_r,
\qquad
m_{r,t}=\left|\{i\le t : s_i\ge r\}\right|.
\]

\begin{table}[t]
    \centering
    \small
    \setlength{\tabcolsep}{4pt}
    \caption{\textbf{Idealized per-token costs.}
    Add $F_s$ to obtain total matrix-product FLOPs; multiply the KV columns
    by $2bN$ to obtain bytes. Reads count logical accesses, not physical
    HBM traffic. Copying aliases existing entries rather than storing
    duplicates.}
    \label{tab:kv-policy-costs}

    \begin{tabular}{@{}lccc@{}}
        \toprule
        Policy
        & FLOPs beyond $F_s$
        & KV reads$/\,(2bN)$
        & KV writes$/\,(2bN)$ \\
        \midrule

        Backfilling
        & $4NtD_s+\Delta F_t$
        & $tU_s+\sum_{(i,r)\in\mathcal B_t} i v_r$
        & $U_s+\sum_{(i,r)\in\mathcal B_t} v_r$ \\

        Masking
        & $4N\sum_{r\le s} d_r m_{r,t}$
        & $\sum_{r\le s} v_r m_{r,t}$
        & $U_s$ \\

        Copying
        & $4NtD_s$
        & $tU_s$
        & $U_s$ \\

        Imputing
        & $4NtD_s+4ND_s(U_K-U_s)$
        & $tU_s$
        & $U_K$ \\

        \bottomrule
    \end{tabular}
\end{table}

For backfilling, $a_i$ is token $i$'s largest cached exact budget:
\[
\mathcal B_t
=
\{(i,r): i<t,\ a_i<r\le s\},
\qquad
\Delta F_t
=
\sum_{(i,r)\in\mathcal B_t}
\left(C_r+4Ni d_r\right).
\]

Reconstruction assumes retained intermediate activations and requires no historical readouts. Counts use two FLOPs per multiply--add and omit elementwise work, routing, and non-KV traffic; physical traffic and runtime depend on the implementation.

\subsection{Mamba Block Formulation and State Reuse}
\label{app:mamba-block}

Let
the expanded channels be partitioned into groups of complete recurrent heads $\mathcal H_1,\ldots,\mathcal H_K$, with fixed head dimension $d_h$ and state dimension $N$. Budget $s$ activates $\bigcup_{r\le s}\mathcal H_r$. Write $u_t$ for the prefix-invariant, pre-normalized residual stream at token $t$. For $h\in\mathcal H_r$, triangular projections of $u_{t,\le r}$ produce a gate $z_{h,t}$, recurrent input $x_{h,t}\in\mathbb R^{d_h}$, and step input $\delta_{h,t}$. The projections for the shared vectors $B_t,C_t\in\mathbb R^N$ read only $u_{t,\le1}$; unlike the head channels, their state dimensions are retained in full at every budget.

Let $\operatorname{DWConv}$ be a causal depthwise convolution applied separately to $x$, $B$, and $C$, and let $\phi=\operatorname{SiLU}$. The scan inputs and head-specific coefficients are
\begin{equation}
\begin{aligned}
\bar x_{h,t}&=\phi\!\left(\operatorname{DWConv}_h(x_h)_t\right),
&\bar B_t&=\phi\!\left(\operatorname{DWConv}_B(B)_t\right),
&\bar C_t&=\phi\!\left(\operatorname{DWConv}_C(C)_t\right),\\
\Delta_{h,t}&=\operatorname{softplus}(\delta_{h,t}+b_h),
& A_h&=-\exp(a_h),
& D_h&\in\mathbb R.
\end{aligned}
\label{eq:triod-mamba-inputs}
\end{equation}
Here $a_h,b_h,D_h$ are head-specific parameters. From an initial state $S_{h,0}\in\mathbb R^{d_h\times N}$, each active head follows \cref{eq:triod-mamba-scan}. The gate acts before output normalization: concatenate the head outputs and gates in head order and set $v_t=o_t\odot\phi(z_t)$. If $M_r$ is the number of expanded channels through group $r$, hierarchical RMS normalization and the block-triangular output projection give, for channel $i$ in group $r$,
\begin{equation}
\widehat v_{t,i}=\gamma_i
\frac{v_{t,i}}{\sqrt{M_r^{-1}\sum_{j\le M_r}v_{t,j}^{2}+\epsilon}},
\qquad
m_{t,r}=\sum_{q\le r}W^O_{rq}\widehat v_{t,q},
\qquad h'_{t,r}=h_{t,r}+m_{t,r}.
\label{eq:triod-mamba-output}
\end{equation}

For any budgets $s\le s'$, consider a head $h\in\mathcal H_r$ with $r\le s$. Its projected $x,z,\delta$ agree because $u_{t,\le r}$ agrees by prefix invariance; its shared $B,C$ agree because $u_{t,\le1}$ agrees. Depthwise convolution uses only that head's channels or the shared $B,C$ channels, so its convolution history and barred inputs also agree by induction on tokens. Starting from identical recurrent states, \cref{eq:triod-mamba-scan} then gives identical $S_{h,t}$ and $o_{h,t}$ at every token. At masked positions, neither the convolution histories nor the recurrent states advance. Finally, the gate and the fixed-boundary statistics in \cref{eq:triod-mamba-output} leave earlier normalized groups unchanged, while its triangular projection prevents later heads from modifying earlier residual blocks. Thus the complete Mamba residual block is prefix invariant for the same token history and padding mask. On a width increase, existing convolution and recurrent states remain valid. Exact escalation must backfill the missing residual features and newly activated heads' convolution and recurrent states over earlier tokens, proceeding layer by layer because deeper mixers depend on the new outputs of earlier layers. \cref{tab:mtbench_sequence_lengths} reports how different policies can affect sequence length.

\begin{table}[t]
\centering
\small
\begin{tabular}{lr}
\hline
\textbf{Method} & \textbf{Mean generated tokens} \\
\hline
Masking     & 228.2 \\
Copying     & 255.0 \\
Imputing    & 225.3 \\
Backfilling & 209.2 \\
Vanilla     & 207.2 \\
\hline
\end{tabular}
\caption{\textbf{Average generated sequence lengths on MT-Bench.}
Results use 90B-token pretraining followed by instruction tuning.
Lengths exclude prompt tokens.
Each KV strategy averages over all nine evaluated routing settings,
80 questions, and two generation seeds (1,440 responses);
Vanilla averages over 80 questions and two seeds (160 responses).
Generation is capped at 1,024 tokens.}
\label{tab:mtbench_sequence_lengths}
\end{table}

\section{Proofs}
\label{sec:appendix:proofs}

\subsection{Deep linear networks and truncated SVD}
\label{app:deep-linear-svd}

Throughout, $A^\star=U\Sigma V^\top$ is a compact SVD with $\Sigma=\operatorname{diag}(\sigma_1,\ldots,\sigma_k)$, $\sigma_1\ge\cdots\ge\sigma_k>0$, and $A_r=U\Sigma M_rV^\top$ is its rank-$r$ truncation, where $M_r$ is the prefix mask of \cref{sec:preliminaries} with single-coordinate blocks. Equation
\begin{equation}
\label{eq:svd-prefix}
    A_rx=U\,M_r\,(\Sigma V^\top x)
    =\sum_{i=1}^{r}u_i\bigl(\sigma_iv_i^\top x\bigr),
\end{equation}
shows that a network with dense input map $\Sigma V^\top$, one hidden layer of width $k$, and dense readout $U$ realizes every $A_r$ as a prefix. The question is whether inserting hidden layers, which must be block-triangular to preserve prefix invariance, restricts this. \Cref{thm:deep-linear-svd} states that it does not: at any depth, and for any choice of the hidden maps but one, the network still computes $A^\star$ at full width and $A_r$ at every deployed prefix.

For deployed budgets $\mathcal K=\{r_1,\ldots,r_J\}$ with $r_J=k$, define the block-triangular class
\begin{equation}
\mathcal L_{\mathcal K}=\bigl\{G\in\mathbb R^{k\times k}:\ M_rG(I-M_r)=0\ \text{for all } r\in\mathcal K\bigr\}.
\label{eq:triangular-class}
\end{equation}

\begin{lemma}[Closure]
\label{lem:closure}
$\mathcal L_{\mathcal K}$ is closed under products and under inversion of its invertible elements, contains every diagonal matrix, and satisfies $M_rG=M_rGM_r$ for all $G\in\mathcal L_{\mathcal K}$ and $r\in\mathcal K$.
\end{lemma}
\begin{proof}
The last identity restates \eqref{eq:triangular-class}. For products, $M_rGH(I-M_r)=M_rGM_rH(I-M_r)+M_rG(I-M_r)H(I-M_r)=0$, since $M_rH(I-M_r)=0$ and $M_rG(I-M_r)=0$. For inverses, fix $r\in\mathcal K$ and write $G$ in block form with respect to $\mathbb R^k=\mathbb R^r\oplus\mathbb R^{k-r}$; then $G=\left[\begin{smallmatrix}G_{11}&0\\G_{21}&G_{22}\end{smallmatrix}\right]$ with $G_{11},G_{22}$ invertible because $\det G=\det G_{11}\det G_{22}$, and $G^{-1}=\left[\begin{smallmatrix}G_{11}^{-1}&0\\-G_{22}^{-1}G_{21}G_{11}^{-1}&G_{22}^{-1}\end{smallmatrix}\right]$ has the same zero block. Diagonal matrices satisfy \eqref{eq:triangular-class} trivially.
\end{proof}

\begin{proof}[Proof of \cref{thm:deep-linear-svd}]
We first treat hidden widths $d_1=\cdots=d_L=k$. For $L=1$ take $G_1=\Sigma$, recovering the factorization in \eqref{eq:svd-prefix}. For $L\ge2$, choose arbitrary invertible $G_1,\ldots,G_{L-1}\in\mathcal L_{\mathcal K}$ and set $G_L=\Sigma(G_{L-1}\cdots G_1)^{-1}$, which lies in $\mathcal L_{\mathcal K}$ by \cref{lem:closure}. The network is $x\mapsto UG_L\cdots G_1V^\top x$ with dense endpoints $V^\top$ and $U$, and its full-width product is $U\Sigma V^\top=A^\star$.

For the prefix claim, the width-$r$ submodel retains the first $r$ coordinates after every hidden map and reads out through the first $r$ columns of $U$, i.e., computes $UM_rG_LM_rG_{L-1}\cdots M_rG_1V^\top x$. By \cref{lem:closure}, $M_rG_LM_r=M_rG_L$, so $M_rG_LM_rG_{L-1}=M_r(G_LG_{L-1})$; since products stay in $\mathcal L_{\mathcal K}$, repeating the step gives $M_rG_L\cdots G_1=M_r\Sigma=\Sigma M_r$. Hence the prefix computes $U\Sigma M_rV^\top x=A_rx$ for every $r\in\mathcal K$. The same identity shows that feeding $G_1$ the full $V^\top x$ or its prefix $M_rV^\top x$ gives the same result, so the dense input map need not be masked.

For general widths $d_\ell\ge k$, let $E_\ell\in\mathbb R^{d_\ell\times k}$ embed $\mathbb R^k$ into the first $k$ coordinates of layer $\ell$, and define $\widetilde W_1=E_1G_1V^\top$, $\widetilde W_\ell=E_\ell G_\ell E_{\ell-1}^\top$ for $2\le\ell\le L$, and readout $UE_L^\top$. Each $\widetilde W_\ell$ places $G_\ell$ in its top-left $k\times k$ block and is zero elsewhere, hence block-triangular for every boundary $r\le k$. Since $E_\ell^\top E_\ell=I_k$, the unmasked product is $UG_L\cdots G_1V^\top=A^\star$. Writing $M_r^{(\ell)}$ for the prefix mask on layer $\ell$, we have $M_r^{(\ell)}E_\ell=E_\ell M_r$ and $E_\ell^\top M_r^{(\ell)}=M_rE_\ell^\top$, so the masked product reduces to the width-$k$ case and equals $A_r$.
\end{proof}

The theorem shows that depth is free in the linear case: the dense endpoints fix the singular basis, and within that basis any invertible block-triangular trunk can be compensated by one further block-triangular factor. The simplest instance takes $G_1=\cdots=G_{L-1}=I$ and $G_L=\Sigma$, or symmetrically $\Sigma^{1/2}$ on each endpoint with identity hidden maps, so that each hidden coordinate carries one singular direction and widening appends $\sigma_{r+1}u_{r+1}v_{r+1}^\top x$ without altering the rest.

\begin{lemma}[Squared loss with general input covariance]
\label{lem:whitening}
Let $\mathcal R(M)=\mathbb E\|y-Mx\|_2^2$ with $C_x=\mathbb E[xx^\top]\succ0$, and let $A^\star$ minimize $\mathcal R$. Then for every $M$,
$\mathcal R(M)-\mathcal R(A^\star)=\|(M-A^\star)C_x^{1/2}\|_F^2$. Consequently, if $A^\star C_x^{1/2}=U\Sigma V^\top$, the minimizer of $\mathcal R$ over rank $\le r$ is $A_rC_x^{-1/2}$ with excess risk $\sum_{i>r}\sigma_i^2$.
\end{lemma}
\begin{proof}
The normal equations give $\mathbb E[(y-A^\star x)x^\top]=0$, so $\mathcal R(M)=\mathcal R(A^\star)+\operatorname{tr}\bigl((M-A^\star)C_x(M-A^\star)^\top\bigr)$, which is the stated identity. Since $C_x^{1/2}$ is invertible, $\operatorname{rank}(MC_x^{1/2})=\operatorname{rank}(M)$, so minimizing over rank $\le r$ is the best rank-$r$ approximation of $A^\star C_x^{1/2}$, given by Eckart--Young.
\end{proof}
Applying \cref{thm:deep-linear-svd} to $A^\star C_x^{1/2}$ and composing the input map with $C_x^{-1/2}$, which remains dense, yields a prefix-invariant deep linear network whose width-$r$ prefix attains the optimal excess risk at every deployed $r$; this is the risk statement following the theorem in the main text.

\section{Experimental Setting}
\label{sec:appendix:experiments}

\subsection{Training Details}
\label{sec:appendix:experiments:training}
For \llama{} and \nemotron{} we train on a data mixture inspired by the one proposed by \cite{allal2025smollm}, scaled down and adapted according author recommendations from $11$T to $30$B tokens, comprising 60\% FineWeb-Edu, 20\% Stack-Edu, 15\% FineMath-4+, and 5\% Cosmopedia-v2.
For \method{}, unless otherwise specified, we train using Hierarchical Self-Distillation (\hsd{}) and Membership-Normalized Gradients (\mng{}), and choose the same nested widths both for \method{} and Slimmable networks. For \llama{} we choose the nested widths corresponding to using $3$,$6$ and $8$ over the total $8$ attention/KV cache heads; for \nemotron{} and \qwen{} we train nested widths corresponding to all of $4$ heads.

Over all the main experiments, we fix a sequence length of $2048$ and global batch size of $1024$, and use AdamW optimizer with standard parameters and learning $\eta$ rate separately tuned for each method. We found the best ones being consistent across the methods, being $\eta=5e-5$ for both architectures. We employ a Warmup-Schedule-Decay (WSD) schedule with a warmup of $500$ steps and a linear cooldown phase to $0$ in the last $10\%$ of training, following the practice of \cite{allal2025smollm}.

\subsubsection{Instruction Tuning Details}
\label{sec:appendix:experiments:instruction_tuning}
\begin{revised}
We instruction-tune the 90B-token-pretrained Llama model on \texttt{yahma/alpaca-cleaned} for two epochs (14M total response tokens) \citep{alpaca}. All parameters are updated using response-only cross-entropy averaged equally across widths 0.375, 0.75, and 1.0. We use AdamW with a peak learning rate of 2e-5, cosine decay, 1\% warmup, weight decay 0.1, and an effective batch size of 64. Exact duplicate examples are removed, as are examples exceeding 4,096 tokens.
\end{revised}

\subsection{Evaluation Details}
\label{sec:appendix:experiments:evaluation}

\subsubsection{Benchmarks}
\label{sec:appendix:experiments:benchmarks}
We evaluate zero-shot multiple-choice accuracy on ARC-Challenge, ARC-Easy,
HellaSwag, OpenBookQA, PIQA, and WinoGrande using the lm-evaluation-harness
\citep{eval-harness}. We score each answer choice by full-history teacher-forced
log-likelihood and report the native \texttt{acc\_norm} metric, except for
WinoGrande, which uses \texttt{acc}. Scores and per-task mean decode FLOPs/token
are averaged equally across the six tasks, rather than pooling questions or
tokens across tasks. FLOPs are analytical estimates, not measured latency.

\textbf{Evaluation splits.}
\Cref{fig:cascading_benchmarks} uses the harness's default evaluation splits:
test for ARC-Challenge, ARC-Easy, and OpenBookQA, and validation for the other
three tasks. The router-training-data ablation, its per-dataset breakdown, and
the depth-based comparison
(\cref{fig:router-training-data,fig:cascading-per-dataset,app:depth-cascades})
instead use validation splits for all six tasks. Every evaluation uses the
complete selected split.

\textbf{Cascading policies.}
Unless otherwise specified, each language model is frozen and its KL policy is
fitted on 3,000 FineWeb-Edu documents, with 1,000 disjoint documents for
calibration, following Sec.~\ref{router}. The resulting policy and threshold
vectors are reused across benchmarks without task-specific tuning. The
task-trained ablation in \cref{abb:router-dataset} replaces these documents
with task-specific training data.

\subsubsection{Generation Quality Evaluation}
We evaluate on the first turn of all 80 MT-Bench questions, generating two responses per question with different random seeds.
We use a temperature of 1.0 across all categories and a maximum of 1,024 generated tokens.
Each response is scored independently on a 1-10 scale by GPT-6-Luna with low reasoning effort, using FastChat's MT-Bench single-answer judging prompts (\texttt{single-v1} and \texttt{single-math-v1}) \citep{Bai_2024}.

\subsubsection{Generation Speedup Measurements and Comparisons}
\label{sec:appendix:experiments:generation_speed}

\textbf{Parameter-matched reference.}
As a timing baseline, we compare \method{} against a parameter-matched dense model, which has the same architecture as Llama-1B aside from the following modifications: hidden width of 1,664 instead of 2,048, FFN intermediate width of 7,552 instead of 8,192, and 24 query heads and 6 KV heads instead of 32 and 8, respectively. It retains 16 layers and a head dimension of 64. Its total parameter count is 918.9M (down from 1.236B), which is within $0.3\%$ of \method{}'s effective 916.5M.

\textbf{Sparsity-matched reference.}
As an additional timing baseline, we apply $37.5\%$ unstructured sparsity to the dense Llama-1B model's attention and FFN projection matrices, using randomly selected zero locations within each row. Embeddings and the output head remain unchanged. We execute this model within vLLM using native PyTorch CSR sparse matrix operations with CUDA graphs enabled for decoding.

\textbf{Measuring speed.}
We measure batch-1 decoding latency on an NVIDIA A100, using our optimized inference engine for \method{} and vLLM for the parameter-matched dense baseline, both in FP16. We time 127 decode steps per prompt, excluding prefill, reporting the median aggregate throughput over five repetitions as milliseconds per token. The dense baseline uses untrained weights and serves only as a timing control.

\subsection{Hierarchical distillation on nested low-rank regression}
\label{app:hsd_vs_sd}

We study a neural network with one hidden layer of 12 neurons, with trainable linear input and output layers and no activations or biases. Each submodel uses a prefix of the hidden neurons, yielding 12 nested predictors with shared parameters. Inputs are sampled independently and uniformly from the unit sphere in 12 dimensions. Targets are generated by a fixed diagonal linear map whose singular values are equally spaced from 1.5 to 0.5, without added observation noise. Under this isotropic input distribution, the optimal population predictor at each width is the corresponding truncated SVD of the target matrix. These optima can be represented simultaneously by prefixes of a single network.

We compare self-distillation from the current full model (SD) with hierarchical self-distillation from the next larger submodel (HSD). Teacher predictions are detached during gradient computation, whereas the full model is supervised with the true targets in both methods. For this regression experiment, distillation uses half-squared Euclidean prediction error, and the training objective averages the losses equally across all widths. We optimize both factor matrices using plain SGD with minibatch size 8 and a constant learning rate. Initial parameter entries are sampled independently from a zero-mean Gaussian distribution with standard deviation 0.12.

We select a separate learning rate for each method via a powers-of-two grid search, minimizing the mean final population suboptimality after 5,000 updates across three tuning seeds. The grid is extended whenever the minimizing candidate lies at a boundary. We then evaluate the methods on five additional seeds, with matched initial parameters and input streams across methods. Figure~\ref{fig:hkd_vs_kd} reports the comparison over 5,000 SGD updates.

Population suboptimality is computed exactly as the population half-squared prediction loss above the corresponding truncated-SVD optimum, averaged over all widths. 
Curves show means over the five evaluation seeds, and shaded regions indicate one standard error.

\subsection{Implementing Efficient Cascading Inference for \method{}}
\label{app:throughput}

We engineer an optimized \method{} inference implementation that reuses intermediate computations across cascade stages, uses stage-specific packed weights and fused kernels, and reduces dispatch overhead through CUDA Graphs with GPU-side conditional execution. Fixed-width baselines receive the same applicable engine optimizations, allowing us to assess the additional benefit of progressive inference. Measurements use FP16 weights and activations with FP32 accumulation for numerically sensitive operations.

\textbf{Numerical precision.}  Implementation uses FP16 for storage and FP32 for accumulation of logits and attention. The use of FP16 over BF16 for cascaded inference is important because intermediate computations face numerical issues during accumulation of intermediate accumulations; accumulations over a subset of channel widths may round differently than an accumulation over the entire width of the model. We observed that BF16 exacerbates these issues.

\begin{revised}
\subsection{Additional Ablations}

\subsubsection{Router training data}
\label{abb:router-dataset}
We compare FineWeb-Edu-trained and per-task-trained policies, keeping
models frozen and policy architectures and optimizer searches unchanged.
Task-specific policies use up to 3,000 fitting and 1,000 calibration questions
from disjoint training subsets (80/20 for smaller datasets). Both settings follow
the benchmark protocol in \cref{sec:appendix:experiments:benchmarks}.
Task-trained and FineWeb-Edu-trained policies yield very similar average
accuracy--FLOPs curves (\cref{fig:router-training-data}).
We therefore use FineWeb-Edu policies across tasks, avoiding task-specific
retraining.

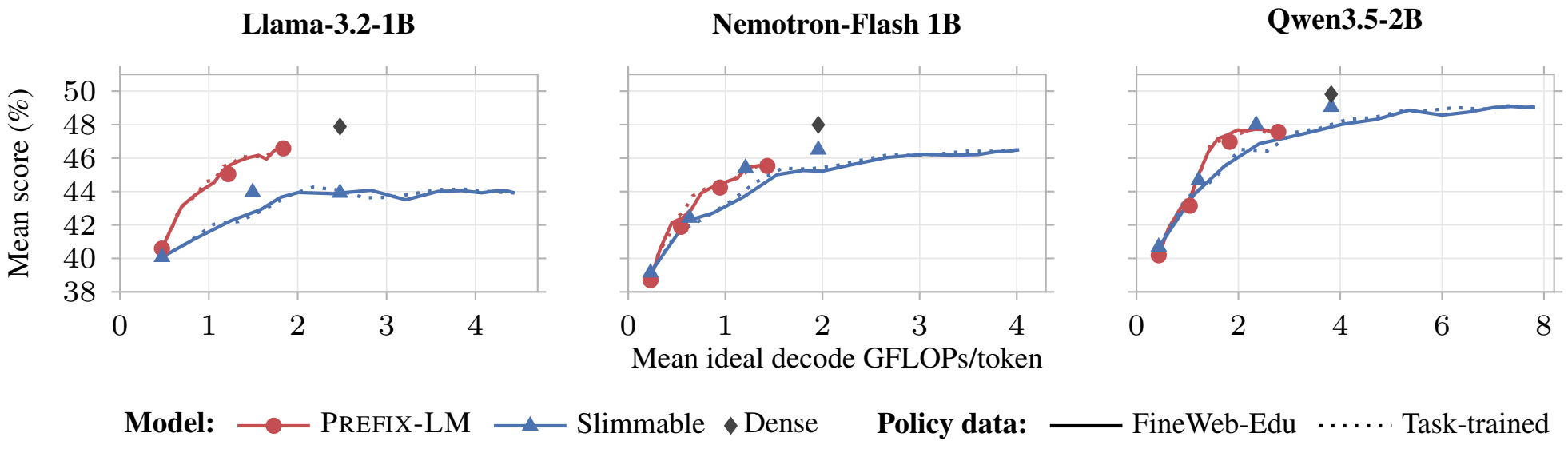
\begin{figure}[htbp]
\centering
\definecolor{routerDataMethod}{HTML}{C44E52}
\definecolor{routerDataSlimmable}{HTML}{4C72B0}
\definecolor{routerDataDense}{HTML}{444444}
\begin{tikzpicture}
\begin{groupplot}[
  paper triple,
  group style={group size=3 by 1,horizontal sep=\PlotGap,y descriptions at=edge left},
  ymin=38,ymax=51,ytick={38,40,42,44,46,48,50},
  scaled ticks=false,
]
\nextgroupplot[title={Llama-3.2-1B},
  xmin=0,xmax=4.7,xtick={0,1,2,3,4},
  ylabel={Mean score (\%)},legend to name=routerDataLegend,legend columns=-1,
  legend style={/tikz/every even column/.append style={column sep=0.1cm}}]
\addlegendimage{empty legend}
\addlegendentry{\textbf{Model:}}
\addlegendimage{routerDataMethod,solid,mark=*}
\addlegendentry{\method{}}
\addlegendimage{routerDataSlimmable,solid,mark=triangle*}
\addlegendentry{Slimmable}
\addlegendimage{routerDataDense,only marks,mark=diamond*}
\addlegendentry{Dense}
\addlegendimage{empty legend}
\addlegendentry{\quad\textbf{Policy data:}}
\addlegendimage{black,solid,no marks}
\addlegendentry{FineWeb-Edu}
\addlegendimage{black,dotted,no marks}
\addlegendentry{Task-trained}
\addplot[routerDataMethod,solid,no marks,forget plot] coordinates {
    (0.472417878,40.588065297) (0.572447450,41.747881998) (0.695786891,43.118849056)
    (0.828861521,43.728265930) (0.951339980,44.183478134) (1.056735848,44.531539254)
    (1.154791190,45.220359025) (1.243107878,45.593410326) (1.344704915,45.839169933)
    (1.450503364,46.034334291) (1.558645281,46.164087033) (1.648128843,45.943339938)
    (1.740228163,46.469274781) (1.802132497,46.562812782) (1.837802897,46.576859698)
  };
\addplot[routerDataMethod,dotted,no marks,forget plot] coordinates {
    (0.472417878,40.588065297) (0.576508246,41.682181541) (0.671717617,43.054925375)
    (0.770793036,43.336614726) (0.869840623,43.905814677) (0.968930854,44.600658325)
    (1.065283649,44.712420309) (1.160797982,45.514878508) (1.258593019,45.631260172)
    (1.355210085,46.089272417) (1.451726885,46.059310670) (1.549313707,46.096470309)
    (1.645339121,46.118923903) (1.739113991,46.473556248) (1.837802897,46.576859698)
  };
\addplot[routerDataSlimmable,solid,no marks,forget plot] coordinates {
    (0.472417878,40.074398464) (0.834256625,41.145401103) (1.244410567,42.244121680)
    (1.586591170,42.931373876) (1.807707520,43.657094774) (2.008343900,43.943754211)
    (2.395793663,43.875214573) (2.822015789,44.075897255) (3.215550379,43.506049026)
    (3.578079500,44.005822400) (3.849331948,44.055043958) (4.069115061,43.922082047)
    (4.235200092,44.043535075) (4.350924761,44.041931054) (4.440739475,43.920297202)
  };
\addplot[routerDataSlimmable,dotted,no marks,forget plot] coordinates {
    (0.472417878,40.074398464) (0.764592540,40.940990433) (1.053634194,42.038464266)
    (1.334274033,42.205343539) (1.612129340,42.893276626) (1.885894851,43.727348519)
    (2.170676174,44.262296978) (2.454438393,44.089397754) (2.749642807,43.623696191)
    (3.031076633,43.654264182) (3.314079731,43.880417463) (3.594911533,44.117799450)
    (3.881906046,44.137663163) (4.155588954,43.999609370) (4.440739475,43.920297202)
  };
\addplot[routerDataMethod,only marks,mark=*,forget plot] coordinates {
    (0.472417878,40.588065297) (1.218514092,45.041756153) (1.837546385,46.576859698)
  };
\addplot[routerDataSlimmable,only marks,mark=triangle*,forget plot] coordinates {
    (0.472417878,40.074398464) (1.492192428,43.962956438) (2.476129169,43.920297202)
  };
\addplot[routerDataDense,only marks,mark=diamond*,forget plot] coordinates {
    (2.476129169,47.880243303)
  };
\nextgroupplot[title={Nemotron-Flash 1B},
  xmin=0,xmax=4.3,xtick={0,1,2,3,4}]
\addplot[routerDataMethod,solid,no marks,forget plot] coordinates {
    (0.229247347,38.698816102) (0.328037467,40.563099244) (0.449195223,42.137085059)
    (0.554432545,42.413964930) (0.649820212,42.927949850) (0.750033404,43.897726663)
    (0.850898126,44.226158780) (0.945541029,44.436465360) (1.044876670,44.655441066)
    (1.124182622,44.796637671) (1.206005232,45.367844294) (1.281768384,45.492782045)
    (1.343420759,45.546601275) (1.396708125,45.477642850) (1.432360396,45.528603020)
  };
\addplot[routerDataMethod,dotted,no marks,forget plot] coordinates {
    (0.229247347,38.698816102) (0.315148890,40.414121931) (0.402721370,41.175016881)
    (0.488551882,42.370299623) (0.576373143,42.871743787) (0.661185084,43.725167865)
    (0.747609873,44.025225325) (0.833932260,44.276605185) (0.923802690,44.145398733)
    (1.008453514,44.568613470) (1.093529282,44.791227446) (1.177638180,45.027384694)
    (1.263481319,45.334434857) (1.345453772,45.563010334) (1.432360396,45.528603020)
  };
\addplot[routerDataSlimmable,solid,no marks,forget plot] coordinates {
    (0.229247347,39.149256313) (0.600846639,42.235593758) (0.879823164,42.730973483)
    (1.193525746,43.700864213) (1.532622805,45.010755667) (1.799336357,45.258594327)
    (1.996922856,45.213510674) (2.290923934,45.593031387) (2.662491957,46.032051967)
    (3.033112518,46.221615268) (3.346474434,46.175302041) (3.604062411,46.211184634)
    (3.780914605,46.377629596) (3.916152550,46.409556579) (4.024843902,46.496964615)
  };
\addplot[routerDataSlimmable,dotted,no marks,forget plot] coordinates {
    (0.229247347,39.149256313) (0.504401299,41.539560719) (0.772492675,42.372476802)
    (1.042794074,43.372251235) (1.308644895,44.536681897) (1.585240522,45.392647059)
    (1.855845580,45.340393119) (2.128903334,45.537357601) (2.392927634,45.864909826)
    (2.670221436,46.185970638) (2.940648001,46.159683446) (3.216155530,46.247668662)
    (3.481106776,46.413013251) (3.747799509,46.406495583) (4.024843902,46.496964615)
  };
\addplot[routerDataMethod,only marks,mark=*,forget plot] coordinates {
    (0.229247347,38.698816102) (0.543531750,41.890849402) (0.944438361,44.238367181)
    (1.431967180,45.528603020)
  };
\addplot[routerDataSlimmable,only marks,mark=triangle*,forget plot] coordinates {
    (0.229247347,39.149256313) (0.631726822,42.423614206) (1.207450713,45.410995982)
    (1.956419020,46.496964615)
  };
\addplot[routerDataDense,only marks,mark=diamond*,forget plot] coordinates {
    (1.956419020,47.986912055)
  };
\nextgroupplot[title={Qwen3.5-2B},
  xmin=0,xmax=8.2,xtick={0,2,4,6,8}]
\addplot[routerDataMethod,solid,no marks,forget plot] coordinates {
    (0.436367696,40.185463161) (0.632777265,41.823573281) (0.860798637,42.995682213)
    (1.045024531,43.547795231) (1.234242917,45.135550847) (1.408747983,46.364164224)
    (1.598058278,47.156436738) (1.787494543,47.385688833) (1.983589786,47.676535363)
    (2.167287500,47.626113636) (2.330491209,47.724169381) (2.475799894,47.704433535)
    (2.622715122,47.598052695) (2.718352388,47.507879986) (2.785190720,47.563553886)
  };
\addplot[routerDataMethod,dotted,no marks,forget plot] coordinates {
    (0.436367696,40.185463161) (0.608906995,41.627530335) (0.779188440,42.421646946)
    (0.949487794,43.357918324) (1.119042617,44.049999895) (1.286423471,45.449056388)
    (1.452073474,46.794191397) (1.624810136,47.029139632) (1.794161947,47.248405850)
    (1.958763020,47.343825118) (2.125114880,47.598855500) (2.290833833,47.733561495)
    (2.455095961,47.624431155) (2.614193014,47.478138363) (2.785190720,47.563553886)
  };
\addplot[routerDataSlimmable,solid,no marks,forget plot] coordinates {
    (0.436367696,40.683102046) (1.132423379,43.825623964) (1.727016884,45.522911085)
    (2.414729585,46.867395298) (2.991655915,47.251223638) (3.515328946,47.618990891)
    (4.030481033,48.014091601) (4.714142876,48.314753262) (5.355718585,48.860896298)
    (5.996995602,48.565062949) (6.530923016,48.752732081) (6.991888561,49.009765285)
    (7.354555958,49.085307725) (7.627205994,49.032753185) (7.826926881,49.051656054)
  };
\addplot[routerDataSlimmable,dotted,no marks,forget plot] coordinates {
    (0.436367696,40.683102046) (0.965760504,43.580336807) (1.499297796,44.678433368)
    (2.029683987,46.510655595) (2.568528308,46.422549277) (3.092508521,47.523192315)
    (3.611295506,47.792183739) (4.132371586,48.301300378) (4.660117802,48.424555521)
    (5.182936905,48.797879940) (5.730640422,48.840836503) (6.263986460,49.015600387)
    (6.789076507,48.917964551) (7.303546672,49.117492315) (7.826926881,49.051656054)
  };
\addplot[routerDataMethod,only marks,mark=*,forget plot] coordinates {
    (0.436367696,40.185463161) (1.045897888,43.152629876) (1.828590576,46.969552770) (2.784445760,47.563553886)
  };
\addplot[routerDataSlimmable,only marks,mark=triangle*,forget plot] coordinates {
    (0.436367696,40.683102046) (1.219060384,44.670303137) (2.348078064,47.987190615) (3.823420736,49.051656054)
  };
\addplot[routerDataDense,only marks,mark=diamond*,forget plot] coordinates {
    (3.823420736,49.816149012)
  };
\end{groupplot}
\node[anchor=north,font=\PlotFont,inner sep=0pt]
  at ([yshift=-0.45cm]group c2r1.south) {Mean ideal decode GFLOPs/token};
\node[anchor=north,inner sep=0pt]
  at ([yshift=-\PlotLegendGap]group c2r1.south)
  {\pgfplotslegendfromname{routerDataLegend}};
\end{tikzpicture}
\caption{\textbf{Task-specific router training offers little average benefit over FineWeb-Edu.}
Averaged over six tasks, both training sources yield similar accuracy--FLOPs
curves for \method{} and Slimmable. This supports reusing FineWeb-Edu-trained
policies across tasks.}
\label{fig:router-training-data}
\end{figure}

\subsubsection{Architectural and parameter-count costs}
\label{abb:decomp_arch}
\begin{figure}
    \centering
    \includegraphics[width=0.98\linewidth]{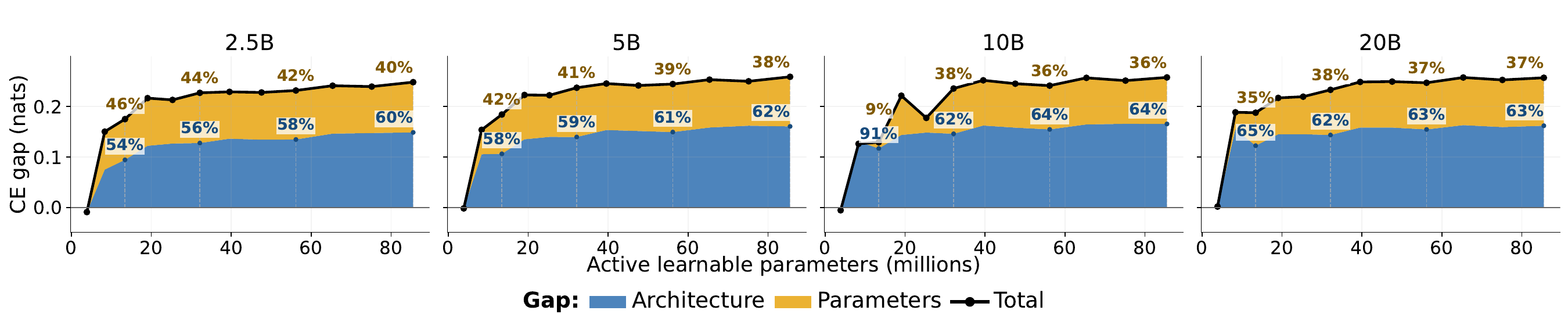}
    \caption{
    \textbf{Independent-model loss-gap decomposition:}
    the decomposition between independently trained \method{} and Slimmable architectures is stable across training budgets.
    }
    \label{fig:app:triag_analysis}
\end{figure}

A natural question is how big the gap between between \method{} and Slimmable architectures is and how much of it depends on the sparsity introduced by triangular weights. Understanding this decomposition provides useful insights to predict how scaling the model size affects its performance.
The plot in \cref{fig:abl:triag_analysis}(a) is obtained by averaging the ones in \cref{fig:app:triag_analysis} over training budgets. Hereafter we provide the details behind the analysis and the extended results, showing that the error gap decomposition is approximately constant across training token budgets. 

This comparison is carried out using independently trained models: the respective architecture are fixed at the one supporting the $12$ nested submodels, but we train them separately, for the total training budget.
In the case of \method{}, it means that we fix the block triangular structure needed to support prefix-invariance with all the submodels, which leaves the model with less active parameters. This is intended to eliminate the confounder of nested model optimization, which would favor \method{} because it optimizes all the model at each step (see \cref{fig:abl:scaling}(c-d)).

To show how the gap decomposes, we compare across each supported width:
\begin{itemize}
    \item $L_T$: the loss of the \method{} architecture. In particular, the model still supports prefix-invariance across all the $12$ submodel, but is trained only to serve one of them. 
    \item $L_S$: the loss of Slimmable architecture, which corresponds to a version of the original architecture that uses a fraction of attention heads and projection dependent of the chosen width
    \item $L_D$: the loss of a parameter-matched dense version of the architecture, obtained by properly resizing the dimensions of attention heads and projections. We found empirically that this $L_D$ and $L_S$ lie in the same parameter-loss front for this study on GPT-2.
\end{itemize}
Given these references, the total gap decomposes as:
\[
\underbrace{L_T-L_S}_{\text{total gap}}
=
\underbrace{L_T-L_D}_{\text{architectural component}}
+
\underbrace{L_D-L_S}_{\text{parameter-count component}}.
\]
The first component measures the remaining disadvantage of \method{} at a matched parameter budget, while the second measures the loss increase associated with reducing the dense model's parameter count to that budget.

\subsubsection{Loss evaluation for complex tasks}
Our main results mainly consider commonsense downstream tasks, while evaluation on math and code, usually regarded as more complex tasks, are not considered. This is not related to the capabilities of our models, but only to practical considerations regarding model training for these tasks. As outlined by \cite{allal2025smollm}, good performance in code and math only emerge already for a traditional non-nested architecture after extended training. In particular, they consider the same \llama{} of our study and pretrain it with a similar recipe for $11$T tokens. roughly $367\times$ the chinchilla training budget allocated in this study. Under that setting, they prove performance similar to the official pretrained checkpoint from Meta.

However, to show that slimmable models and in particular \method{} do not hurt the optimization capabilities for more complex tasks, in \cref{fig:app:llama_codemath_match} we show that the loss scales across widths and training budgets similarly for (a) our held-out test set which resambles the composition of the training data mixture, (b) the external GSM8K dataset for math domain and (c) MBPP for code. In particular, both methods show that under scaled training the full model reaches performances closer to the dense vanilla reference.

\begin{figure}
\begingroup%

\definecolor{pfB25}{HTML}{9cc9e1}%
\definecolor{pfO25}{HTML}{fdad69}%
\definecolor{pfG25}{HTML}{bbbbbb}%
\definecolor{pfB5}{HTML}{519ccc}%
\definecolor{pfO5}{HTML}{f57622}%
\definecolor{pfG5}{HTML}{7f7f7f}%
\definecolor{pfB10}{HTML}{1b69af}%
\definecolor{pfO10}{HTML}{cd4401}%
\definecolor{pfG10}{HTML}{474747}%
\definecolor{pfB20}{HTML}{083573}%
\definecolor{pfO20}{HTML}{852904}%
\definecolor{pfG20}{HTML}{060606}%
\definecolor{pfB30}{HTML}{5ba3d0}%
\definecolor{pfB90}{HTML}{083c7d}%
\definecolor{pfBkey}{HTML}{0d57a1}%
\definecolor{pfO30}{HTML}{f87f2c}%
\definecolor{pfO90}{HTML}{8e2d04}%
\definecolor{pfOkey}{HTML}{b03903}%
\definecolor{pfArchitecture}{HTML}{2166AC}%
\definecolor{pfParameters}{HTML}{E69F00}%

\newcommand{\pfkey}[3]{\tikz[baseline=-.5ex]{%
  \draw[#1,#2,line width=.8pt] (0,0)--(.30,0);%
  \ifcase#3\or%
    \fill[#1] (.15,0) circle[radius=1.3pt];\or%
    \fill[#1] (.115,-.035) rectangle (.185,.035);\or%
    \draw[draw=#1,fill=white,line width=.6pt] (.115,-.035) rectangle (.185,.035);\or%
    \fill[#1] (.15,.05)--(.20,0)--(.15,-.05)--(.10,0)--cycle;%
  \fi}}%
\newcommand{\pfbox}[1]{\tikz[baseline=-.5ex]{\fill[#1,opacity=.8] (0,-.025) rectangle (.30,.075);}}%
\pgfplotsset{pfaxis/.style={paper triple,
  height=1.6cm,
  scaled ticks=false,/pgf/number format/1000 sep={},
  title style={align=center},enlarge x limits=.04}}
\pgfplotstableread[row sep=\\,col sep=space]{%
tx t30 t90 sx s30 s90 \\%
235.36512 2.61836881771 2.51132100247 235.36512 2.70019383339 2.54920667235 \\%
607.569408 2.41670065435 2.30022457141 744.408576 2.43730577128 2.28738579524 \\%
916.523008 2.36980153503 2.24795901576 1235.8144 2.39448729766 2.23377571781 \\%
}\pfMixture%
\pgfplotstableread[row sep=\\,col sep=space]{%
tx t30 t90 sx s30 s90 \\%
235.36512 2.07540563839 1.87519090818 235.36512 2.11806855043 1.91362521203 \\%
607.569408 1.77126198901 1.59712551614 744.408576 1.77745706687 1.55593539679 \\%
916.523008 1.70959739636 1.54707332458 1235.8144 1.7284174173 1.49613443526 \\%
}\pfMath%
\pgfplotstableread[row sep=\\,col sep=space]{%
tx t30 t90 sx s30 s90 \\%
235.36512 1.68806739763 1.59975982514 235.36512 1.70313650855 1.63237157273 \\%
607.569408 1.49793837043 1.42945803717 744.408576 1.51382478917 1.44327937357 \\%
916.523008 1.46979160841 1.38914591573 1235.8144 1.49682043673 1.41846976194 \\%
}\pfCode%

\begin{tikzpicture}
\begin{groupplot}[pfaxis,group style={group size=3 by 1,horizontal sep=\PlotGap},
 xmin=180,xmax=1300,xtick={200,600,1000}]
\nextgroupplot[title={(a) Mixture validation},ymin=2.2,ymax=2.74,ylabel={Held-out cross-entropy $\downarrow$}]
\addplot[color=pfB30,mark=*] table[x=tx,y=t30] {\pfMixture};
\addplot[color=pfB90,mark=*] table[x=tx,y=t90] {\pfMixture};
\addplot[color=pfO30,mark=square*] table[x=sx,y=s30] {\pfMixture};
\addplot[color=pfO90,mark=square*] table[x=sx,y=s90] {\pfMixture};
\addplot[black,dashed,no marks] coordinates {(180,2.25762538635) (1300,2.25762538635)};
\addplot[black,only marks,mark=diamond*] coordinates {(1235.8144,2.25762538635)};
\nextgroupplot[title={(b) Math: GSM8K},ymin=1.47,ymax=2.15]
\addplot[color=pfB30,mark=*] table[x=tx,y=t30] {\pfMath};
\addplot[color=pfB90,mark=*] table[x=tx,y=t90] {\pfMath};
\addplot[color=pfO30,mark=square*] table[x=sx,y=s30] {\pfMath};
\addplot[color=pfO90,mark=square*] table[x=sx,y=s90] {\pfMath};
\addplot[black,dashed,no marks] coordinates {(180,1.53178146119) (1300,1.53178146119)};
\addplot[black,only marks,mark=diamond*] coordinates {(1235.8144,1.53178146119)};
\nextgroupplot[title={(c) Code: MBPP},ymin=1.38,ymax=1.72]
\addplot[color=pfB30,mark=*] table[x=tx,y=t30] {\pfCode};
\addplot[color=pfB90,mark=*] table[x=tx,y=t90] {\pfCode};
\addplot[color=pfO30,mark=square*] table[x=sx,y=s30] {\pfCode};
\addplot[color=pfO90,mark=square*] table[x=sx,y=s90] {\pfCode};
\addplot[black,dashed,no marks] coordinates {(180,1.40203339706) (1300,1.40203339706)};
\addplot[black,only marks,mark=diamond*] coordinates {(1235.8144,1.40203339706)};

\end{groupplot}
\node[font=\PlotFont,anchor=north] at ($(group c1r1.south west)!.5!(group c3r1.south east)+(0,-.45cm)$) {Active learnable parameters (millions)};
\node[font=\PlotFont,anchor=north] at ($(group c1r1.south west)!.5!(group c3r1.south east)+(0,-\PlotLegendGap)$) {%
\textbf{Method:} \pfkey{pfBkey}{solid}{1} \method{}\hspace{.4em}\pfkey{pfOkey}{solid}{2} Slimmable\hspace{.4em}\pfkey{black}{dashed}{4} Dense full $\cdot$ 30B
\quad \textbf{Tokens:} \pfkey{pfG5}{solid}{0} 30B\hspace{.4em}\pfkey{pfG20}{solid}{0} 90B};
\end{tikzpicture}
\endgroup
    \caption{
    \textbf{Optimization on complex tasks for Llama3.2-1B:} both \method{} and Slimmable correctly optimize for tasks in the training mixture usually regarded as more difficult (math and code). This suggest that, on a properly sized training horizon like the one proved by \cite{allal2025smollm}, the elastic models can reach similar downstream performance on these tasks too. The training recipe used in this work is $\approx367\times$ shorten the one found optimal from the authors, hence why even the baseline has no competitive performance on those tasks.
    }
    \label{fig:app:llama_codemath_match}
\end{figure}
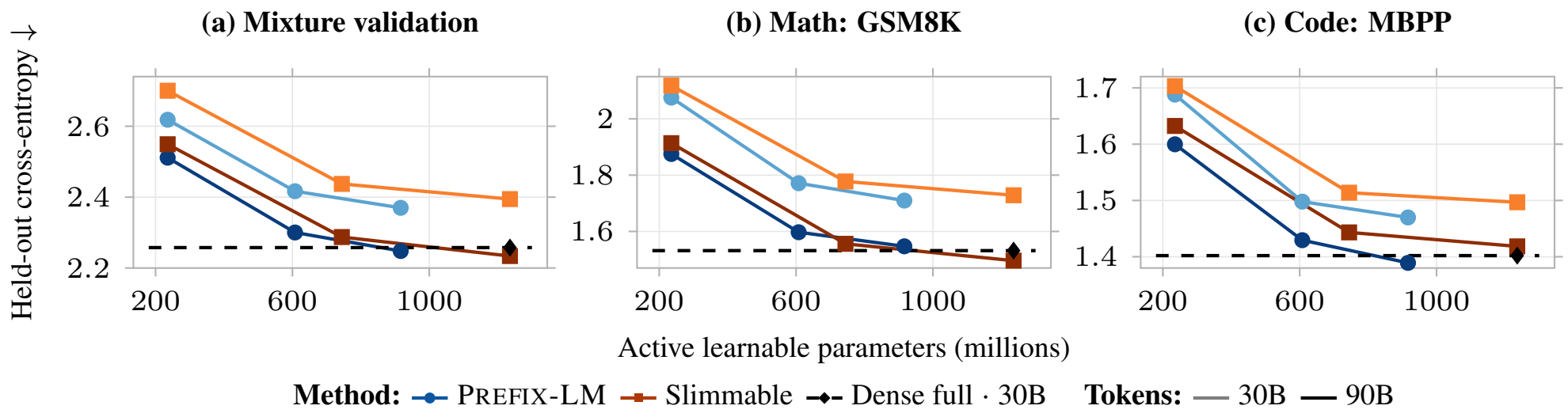

\subsubsection{Replacing KD with HSD}
As in Slimmable Networks, also \method{} can be trained successfully with standard self-distillation from the full model, HSD is not a requirement of the architecture.
Distilling smaller submodels from a larger one is standard for many-in-one models \citep{hinton2015distilling,yu2019slimmable}. HSD changes only the choice of teacher: each submodel learns from the next larger one instead of from the full model. This is motivated by evidence that distillation degrades when the capacity gap between teacher and student is large, and that intermediate teachers help \citep{mirzadeh2019improved}. In slimmable networks, such a chain is impractical because each additional teacher output requires a separate forward pass. Under prefix invariance, all K outputs come from a single pass, so adjacent-teacher distillation is free.
The ablation in \cref{tab:kd_hsd} on \gpt{} replaces HSD with standard KD from the full model, with everything else fixed. The full model reaches a loss of $3.461$ with HSD and $3.494$ with KD ($+0.033$). HSD also has system benefits (\cref{app:telescoping}): in the same setup, it gives 7\% higher training throughput and 15\% lower peak memory than KD.

\begin{table}[h!]
    \centering
        \begin{tabular}{lrrrrr}
        \toprule
         & CE & CE+HSD & CE+KD & KD-HSD & KD excess (\%) \\
        \midrule
        Mean over 12 submodels & 3.969 & 3.863 & 3.867 & 0.005 & 0.12 \\
        Full model   & 3.465          & 3.461 & 3.494 & 0.034 & 0.97 \\
        \bottomrule
        \end{tabular}
    \label{tab:kd_hsd}
    \caption{
    \textbf{HSD provides better distillation than KD:} held-out cross-entropy loss on \gpt{} shows that HSD is a better distillation loss than standard KD.
    }
\end{table}

\clearpage
\subsection{Per-Dataset Cascading Results}
\label{sec:appendix:experiments:per-dataset}
\Cref{fig:cascading-per-dataset} breaks down the router-training-data
comparison in \cref{fig:router-training-data} by dataset for \method{} and
Slimmable. FineWeb-Edu-trained and task-trained policies are evaluated on the
same complete validation splits, with fixed-width submodels and dense
checkpoints as controls, following the protocol in
\cref{sec:appendix:experiments:benchmarks}. Thresholds are fixed before
validation evaluation, and Slimmable cascading inference retains its cumulative FLOP
cost.

\input{figures/cascade_per_dataset}

\clearpage

\subsection{Membership-normalized gradients on nested low-rank regression}
\label{sec:appendix:mng_synthetic}

All language models in this paper are trained with AdamW, which normalizes
each coordinate's update by the running magnitude of its own gradient. A
constant per-block rescaling of the gradient therefore cancels up to the
effect of $\epsilon$, and MNG has little effect in those runs. To isolate
what MNG does, we evaluate it with plain SGD, which has no such invariance.

\paragraph{Setup.} We reuse the nested low-rank regression of
Sec.~\ref{app:hsd_vs_sd} without modification: a one-hidden-layer linear
network with 12 hidden neurons and $K=12$ nested widths, the same data
distribution, initialization, minibatch size, and evaluation protocol. The
only change is the gradient rescaling. Hidden neuron $r$ participates in the
losses of the widths $s \ge r$, so its gradient carries the factor
$c_r = (K-r+1)/K$ of Eq.~\eqref{eq:dilution}; MNG multiplies the gradient of the
parameters of neuron $r$ by $1/c_r$. We compare four configurations: SD and
HSD, each with and without MNG. The learning rate is
selected separately for each configuration with the grid search of
Sec.~\ref{app:hsd_vs_sd}.

\paragraph{Results.} Figure~\ref{fig:mng_synthetic} reports the
suboptimality of each width relative to its truncated-SVD optimum.

\emph{MNG acts where \eqref{eq:dilution} predicts.} At width 1 the factor is
$c_1 = 1$, and with HSD the curves with and without MNG perform similarly. The effect
grows with width, as $c_r$ decreases to $1/K$ for the last neuron.

\emph{With SD, MNG lowers the suboptimality at every width above the
smallest.} The gap is widest at the largest widths, where after 5{,}000
steps the suboptimality is approximately $3\times$ lower with MNG.

\emph{With HSD, MNG accelerates convergence at the larger widths.} At width
12, HSD with MNG reaches a suboptimality of $2\times10^{-3}$ in roughly
700 steps, against 1{,}700 without. Both variants
reach approximately the same value by the end of training.

\emph{HSD converges faster than SD in all cases.} With or without MNG, HSD is
more than an order of magnitude closer to the optimum than SD at every width
after 5{,}000 steps.

\input{figures/mng_hkd_vs_kd_per_width}
\end{revised}

\end{document}